\documentclass[10pt,twocolumn,letterpaper]{article}

\usepackage[pagenumbers]{iccv} % To force page numbers, e.g. for an arXiv version

\definecolor{iccvblue}{rgb}{0.21,0.49,0.74}
\usepackage[pagebackref,breaklinks,colorlinks,allcolors=iccvblue]{hyperref}

\usepackage{marvosym}
\usepackage{algorithm}
\usepackage{algorithmic}
\usepackage{colortbl}
\usepackage{multirow}
\usepackage{makecell}
\usepackage{pifont}
\usepackage{arydshln}

\newcommand\blfootnote[1]{%	
  \begingroup
  \renewcommand\thefootnote{}\footnote{#1}%
  \addtocounter{footnote}{-1}%
  \endgroup
}

\def\paperID{4578} % *** Enter the Paper ID here
\def\confName{ICCV}
\def\confYear{2025}

\definecolor{mygray}{gray}{.92}
\title{EMoTive: Event-guided Trajectory Modeling for 3D Motion Estimation}

\author{
Zengyu Wan,
Wei Zhai,
Yang Cao,
Zhengjun Zha \textsuperscript{\Letter}
\\
USTC\\
{\tt\small wanzengy@mail.ustc.edu.cn}
}

\begin{document}
\twocolumn[{
\maketitle
\begin{center}
    \centering
    \includegraphics[width=.98\linewidth]{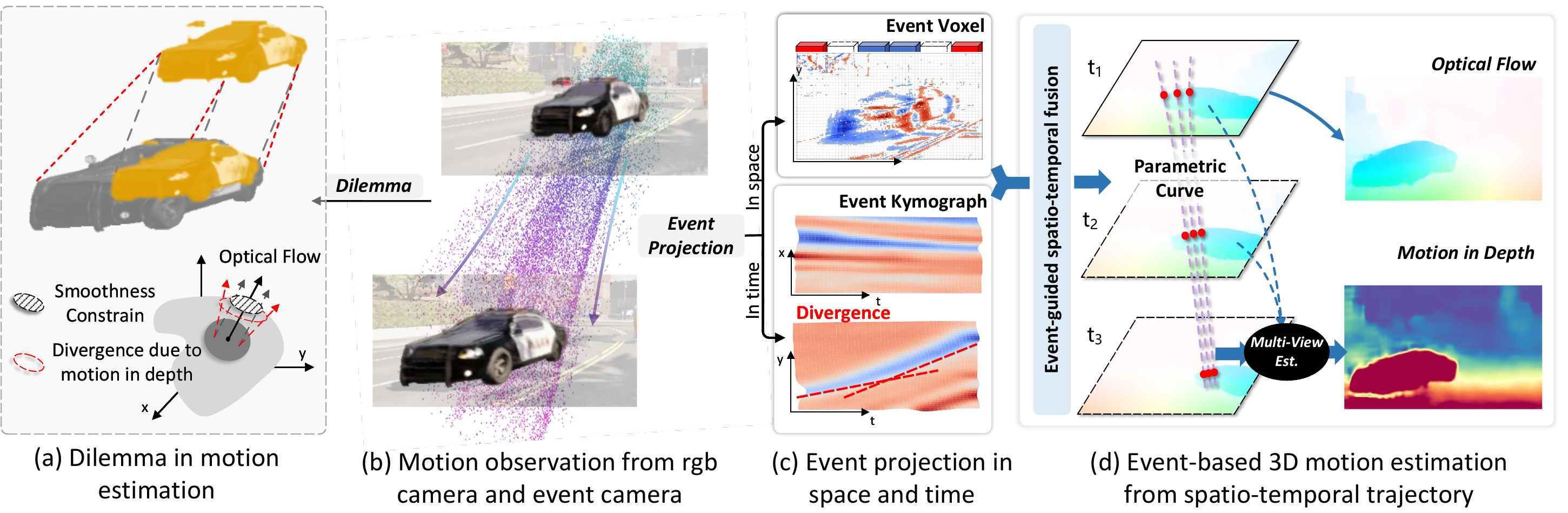}
    \captionof{figure}{(a) The dilemma faced by motion estimation algorithms between the local smoothness assumption and the real divergence due to motion in depth; (b) (c) The spatio-temporal projection of events (Event Kymograph) provides an clear observation of temporal evolution, enabling heterogeneous spatio-temporal motion analysis; (d) 3D motion can be inferred by modelling spatio-temporal trajectories via event-guided non-uniform parametric curves.}
    \label{fig:motivation}
\end{center}
}]

\blfootnote{\Letter: Corresponding author.}

\begin{abstract}
Visual 3D motion estimation aims to infer the motion of 2D pixels in 3D space based on visual cues. The key challenge arises from depth variation induced spatio-temporal motion inconsistencies, disrupting the assumptions of local spatial or temporal motion smoothness in previous motion estimation frameworks. In contrast, event cameras offer new possibilities for 3D motion estimation through continuous adaptive pixel-level responses to scene changes. This paper presents \textbf{EMoTive}, a novel event-based framework that models spatio-temporal trajectories via event-guided non-uniform parametric curves, effectively characterizing locally heterogeneous spatio-temporal motion. Specifically, we first introduce \textbf{Event Kymograph} - an event projection method that leverages a continuous temporal projection kernel and decouples spatial observations to encode fine-grained temporal evolution explicitly. For motion representation, we introduce a density-aware adaptation mechanism to fuse spatial and temporal features under event guidance, coupled with a non-uniform rational curve parameterization framework to adaptively model heterogeneous trajectories. The final 3D motion estimation is achieved through multi-temporal sampling of parametric trajectories, yielding optical flow and depth motion fields. To facilitate evaluation, we introduce \textbf{CarlaEvent3D}, a multi-dynamic synthetic dataset for comprehensive validation. Extensive experiments on both this dataset and a real-world benchmark demonstrate the effectiveness of the proposed method.
\end{abstract}

% The core challenge stems from the depth variation-induced spatio-temporal inconsistent motion patterns, disrupting the local spacial or temporal motion smoothness assumptions from conventional motion estimation frameworks.
% % Previous methods estimate motion under   local space or temporal motion smoothness assumptions yet struggle to address the inherent challenge of spatio-temporal inconsistent motion patterns induced by motion in depth.

% , while jointly preserving spatial motion characteristics
\section{Introduction}
\label{sec:intro }

Visual 3D motion estimation infers the movement of 2D pixels in 3D space, playing a crucial role in spatial intelligence for autonomous systems and enabling diverse applications in dynamic scene understanding \cite{byrne2009expansion, marinho2018guaranteed, badki2021binary, kim2016real}. The key challenge of 3D motion estimation lies in maintaining motion representation consistency throughout the spatial deformation process induced by depth variation. 
% Event cameras, as bio-inspired vision sensors, offer new possibilities for accurate 3D motion estimation with their continuous adaptive pixel-level responses to scene changes 

% The key challenge lies in maintaining motion consistency throughout this deformation process. By their microsecond temporal resolution, high dynamic range, and minimal motion blur, them are promising 

% While conventional approaches employ frame-based cameras or LiDAR for motion perception, their reliance on fixed temporal synchronization limits performance under complex motion patterns and depth variations. The key to 3D motion (or formulated as normalized scene flow) estimation \cite{yang2020upgrading} is maintaining motion consistency throughout the spatial deformation process induced by depth variations.

% In contrast, event cameras - bio-inspired vision sensors that asynchronously report pixel-level brightness changes - offer new opportunities through their microsecond temporal resolution, high dynamic range, and minimal motion blur, making them promising for accurate 3D motion estimation.

% 3D motion (or formulated as normalized scene flow) estimation \cite{yang2020upgrading} can be decomposed into pixel-level planar motion (optical flow) and motion in depth, where depth variations induce spatial  deformation. The key challenge lies in maintaining motion consistency throughout this deformation process.  
% The key challenge of 3D motion estimation lies in maintaining motion consistency throughout the spatial deformation process. 
Existing approaches employ frame-based cameras or LiDAR for 3D motion perception; their reliance on fixed temporal synchronization limits performance under complex motion patterns and depth variations. For example, most optical flow methods \cite{luo2021upflow, liu2021learning, nguyen2022self, yuan2024unsamflow, dong2024memflow} estimate planar motion through inter-frame correspondences of pixels/point clouds, but their insufficient temporal observation necessitates restrictive local spatial motion consistency priors or long temporal consistency to regularize the solution space. Recent efforts \cite{leng2023scaleflow, ling2024scaleflow++, ling2023learning} attempt to decouple planar and depth motion estimation in dual spaces (feature space and scale space), yet the inherent contradiction persists as both spaces originate from the same pixel domain. The root cause stems from heterogeneous spatio-temporal motion: object projections in image space undergo local deformations due to motion in depth, invalidating local temporal invariance assumptions.

Event cameras - bio-inspired vision sensors that asynchronously report pixel-level brightness changes through event streams\cite{gallego2020event, wu2024event} - offer microsecond temporal resolution, high dynamic range, and minimal motion blur, making them promising for accurate 3D motion estimation. 
% Event cameras inherently respond to motion through continuous streams of pixel-level changes, enabling temporal non-stationary modeling. 
Specifically, event streams can form Event Kymograph through x-t and y-t decoupled projections to capture temporal evolution in microseconds, as shown in \cref{fig:motivation}. 
% Event Kymographs provide an explicit representation of temporal changes.
This fine observation enables non-stationary modeling for 3D motion estimation while avoiding the motion dilemma.

% In other words, event cameras can map the local spatial heterogeneity induced by depth variations to temporal variations, thereby resolving the contradiction between spatial motion consistency and depth-induced motion heterogeneity.

% To establish an effective spatiotemporal non-stationary process based on event projections,

Based on the spatio-temporal clues from event cameras,  we propose EMoTive, an event-guided trajectory parameterization framework, to achieve 3D motion estimation. This framework models motion trajectories through non-uniform parametric curves whose deformation characteristics encode 3D motion parameters.  Specifically, it first constructs spatial representations (Event Voxel) through x-y plane projections and temporal representations (Event Kymograph) through x-t and y-t decoupled plane projections. Then, dual cost volumes are constructed on both spatial and temporal representations to guide iterative trajectory refinement. Meanwhile, a density-aware adaptation mechanism is proposed to update the curve parameters by non-uniform sampling of the local event density, which is correlated with the motion strength. The resulting trajectories parameterize pixel motion over time, from which we derive optical flow at arbitrary timestamps and compute depth motion fields through temporal analysis. To enhance depth estimation robustness, we further propose a multi-view fusion algorithm that aggregates observations across adjacent timestamps. To better simulate real-world motion scenarios, we construct the CarlaEvent3D dataset using the CARLA autonomous driving simulator, which includes diverse environments (night, rain, fog, \textit{etc.}) and realistic motions for training and evaluation.

The contributions can be summarized as follows:
\begin{enumerate}
    \item For event-based 3D motion estimation, the \textbf{EMoTive} model is introduced via event-guided non-uniform parametric curve modeling to characterize heterogeneous spatio-temporal motion patterns. 
    \item To encode fine-grained temporal evolution from events, the \textbf{Event Kymograph} is proposed by leveraging a continuous temporal projection kernel from decoupled spatial observations.
    \item During the progress of trajectory formation, a \textbf{Density-aware adaptation} mechanism is designed to guide non-uniform spatio-temporal fusion and curve parameter updating depending on the motion sensitivity of events.
    \item For comprehensive evaluation, the \textbf{CarlaEvent3D} dataset is introduced, which simulates real-world motion under multiple complex scenarios. Extensive experiments on both synthetic and real-world benchmarks reveal the competitive performance of the EMoTive with superior efficiency.
\end{enumerate}

% \begin{enumerate}
%     \item The \textbf{EMoTive} model, an event-based 3D motion estimation architecture, employing event-guided non-uniform parametric curves to model spatio-temporally processes.
%     \item \textbf{Event Kymograph} projection, a temporal representation from decoupled space, leveraging event cameras' low latency to capture spatio-temporal evolution.
%     \item \textbf{Density-aware adaptation} mechanism, utilizing the motion sensitivity of events to guide non-uniform spatio-temporal fusion and curve parameter updating.
%     \item \textbf{CarlaEvent3}D benchmark, a synthetic dataset simulating real-world motion under multi complex scenarios.
% \end{enumerate}
\begin{figure*}[htbp]
    \centering
    \includegraphics[width=\linewidth]{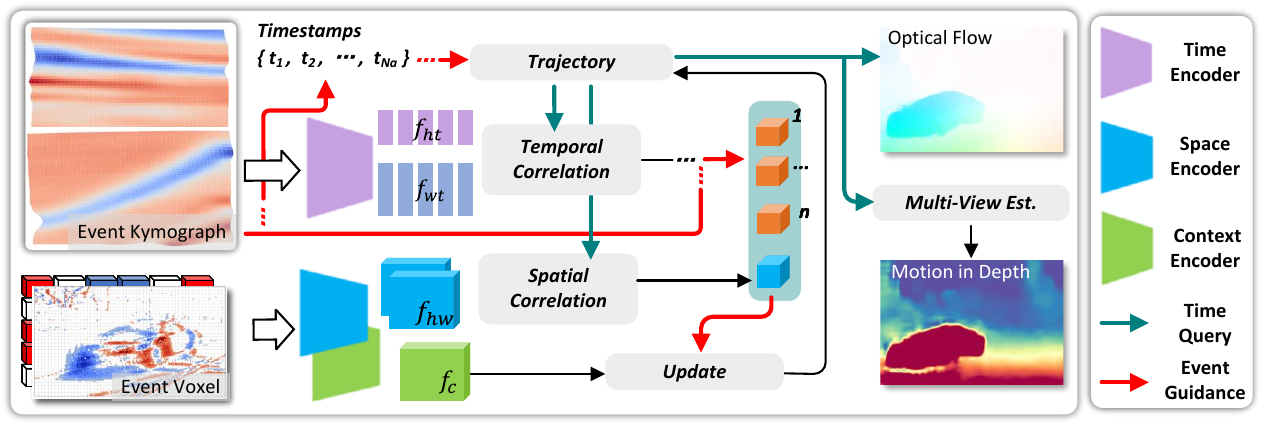}
    \caption{Overall pipeline of EMoTive with event-guided trajectory formation scheme. The Event Voxel and Event Kymograph will be first utilized to construct spatio-temporal dual correlations for motion representation. A density-aware adaptation mechanism is then implemented to fuse spatial and temporal features under event guidance to model trajectories adaptively. The final 3D motion estimation is achieved through multi-temporal sampling of parametric trajectories, yielding optical flow and depth motion field}
    \label{fig:enter-label}
\end{figure*}
\section{Related Work}
\label{sec:related }
\subsection{3D Motion Estimation}
Scene flow was first introduced by Vedula\textit{ et al.} \cite{vedula1999three} to estimate the 3D motion of objects in dense scenes. Various methods have been proposed depending on the available data modalities, such as monocular/stereo images, LiDAR, or RGB-D data. Stereo-based methods  \cite{wedel2008efficient, vcech2011scene, liu2019unsupervised} employ a two-stage pipeline: (1) disparity estimation for 3D reconstruction followed by (2) temporal correspondence matching for scene flow computation. LiDAR-based techniques  \cite{gu2019hplflownet, liu2019meteornet, kittenplon2021flowstep3d, liu2024difflow3d} leverage point cloud registration across frames using geometric matching algorithms. However, these approaches exhibit modality-specific limitations \cite{wan2023rpeflow, zhou2024bring}: Stereo systems require precise extrinsic calibration, while LiDAR suffers from sparse sampling artifacts due to non-uniform beam distribution and surface reflectivity dependencies.

Recent advances in monocular 3D motion estimation have demonstrated promising results through hybrid architectures. Early variational methods \cite{ferstl2014atgv,brickwedde2019mono} jointly optimized motion and depth under restrictive priors \cite{vidal2006two, gotardo2011non}, which limited their generalization capability. A paradigm shift occurred with Yang and Ramanan's normalized scene flow formulation \cite{yang2020upgrading}, enabling depth-agnostic 3D motion estimation through optical expansion analysis. Subsequent works by Han \textit{et al.} \cite{ling2023learning, ling2022scale, ling2024scaleflow++} introduced cross-scale feature fusion mechanisms, unifying optical flow estimation and depth motion prediction in an end-to-end framework. Nevertheless, these image-based methods remain susceptible to spatial deformation artifacts.

\subsection{Event-Based Motion Estimation}
Event cameras offer unique advantages for motion estimation through low latency (µs-level) and  high dynamic range ($>120 dB$).  Early solutions on event-based motion estimation \cite{cook2011interacting, horn1981determining, lucas1981iterative} employed physical constraints like intensity constancy and local motion smoothness. Plane fitting techniques \cite{benosman2013event,aung2018event,low2020sofea} estimate normal flow under local smooth motion assumptions, while contrast maximization frameworks \cite{gallego2018unifying,nagata2021optical,shiba2022secrets,huang2023progressive,han2024event,wan2024event} optimize event motion through spatio-temporal alignment analysis. However, these methods face fundamental limitations in handling complex motion patterns and avoiding local optima. Another line of research pursues learning-based approaches, with architectures like EV-FlowNet \cite{zhu2018ev} and E-RAFT \cite{gehrig2021raft} demonstrating robust 2D optical flow estimation. Recent multi-modal approaches \cite{wan2023rpeflow, zhou2024bring} integrate events with RGB and radar data, exploiting event-driven temporal cues to enhance 3D motion modeling in high-dynamic-range scenarios. Notably, Li \textit{et al.} \cite{li2024blinkvision} developed BlinkVision, a synthetic dataset enabling event-based scene flow learning through novel mesh processing pipelines. Despite progress, the challenge of locally heterogeneous spatio-temporal motion modeling persists.

\begin{figure}[tp]
    \centering
    \includegraphics[width=.97\linewidth]{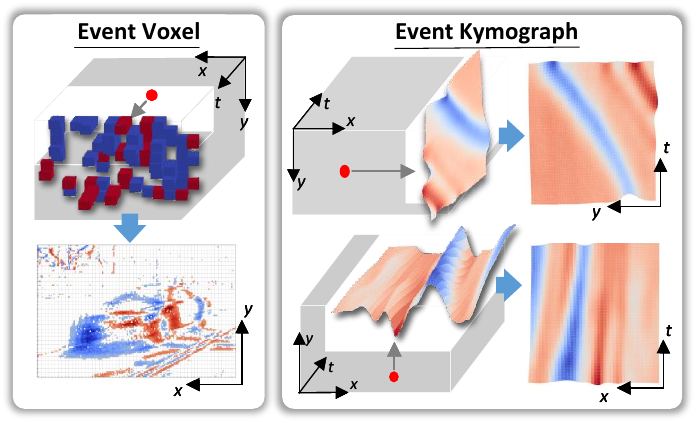}
    \caption{Projection axes of Event Voxel and Event Kymograph. The Event Kymograph captures fine temporal cues in decoupled spatial axes.}
    \label{fig:enter-label}
\end{figure}

\section{Method}
\label{sec:method}
% This paper aims to estimate 3D motion parameters by modeling heterogeneous spatio-temporal motion dynamics through non-uniform parametric curve representations of motion trajectories $\mathcal{T}$
To address the challenge of 3D motion estimation, this paper proposes \textbf{EMoTive}, which models heterogeneous spatio-temporal motion process through non-uniform parametric curve representations of spatio-temporal trajectories $\mathcal{T}$. Specifically, we introduce \textbf{Event Kymograph}, a temporal representation from decoupled space, to leverage temporal correlations from event streams explicitly. The proposed spatio-temporal trajectory $\mathcal{T}$, guided by event streams, is formally defined as: $\mathcal{T}(t, x, y) \ \mapsto \ \mathbb{R}^2$.
% \begin{align}
%     \mathcal{T}(t, x, y) \ \mapsto \ \mathbb{R}^2.
% \end{align}
This continuous representation describes pixel-level motion evolution throughout temporal sequences. Crucially, the trajectory formulation enables the acquisition of pixel motion in image space at arbitrary timestamps, yielding optical flow $\mathcal{O}$. Furthermore, by integrating temporal derivatives, we can establish the motion in depth $\mathcal{M}$, providing comprehensive 3D motion estimation for dynamic scenes.

\subsection{Event Representation}
Asynchronous event streams pose significant challenges for direct analysis and computation. A prevalent solution involves spatio-temporal event aggregation through  planar projections, noted as Event Voxel \cite{deng2022voxel}. This approach aggregates events through temporal quantized, combined with spatio-temporal triangular kernels, formally expressed as:
\begin{gather}
    V(x, y, b) = \sum_i p_i k(x - x_i) k(y - y_i) k(b - t^*_i),
\end{gather}
where the event streams ${(x_i, y_i, t_i, p_i)}_{i\in[1, N]}$  is temporally quantized with bin duration $T_B$: $t^*_i = (t_i - t_1)\, /\, T_B$, and $k(a)$ denotes the triangular sampling kernel:$k(a) = max(0, 1 - |a|)$.  This formulation yields a voxelized tensor $V:\mathbb{R}^{B \times H \times W}$ through spatio-temporal aggregation.

However, conventional event voxel encoding inevitably discards fine cues due to coarse temporal quantization. To overcome these limitations, we propose Event Kymograph projection – a novel spatio-temporal mapping that preserves temporal fidelity through the $x-t, y-t$ spacial axis decoupled projection, formulated as:
% However, event voxel encoding discards redundant time-domain information through time compression and loses some critical fine-grained temporal observation cues. Its spatial encoding also isn’t very convenient for spatiotemporal observation and requires specific design for processing structures. To address this, we adopt the Event-based Kymograph projection form, which compresses space while retaining as much time-domain redundant signals as possible through the $x-t, y-t$ spatiotemporal mapping, providing ample observation perspectives for temporal variation. The specific Event-based Kymograph projection form is as follows:
\begin{gather}
    % \K_x = \sum_i p_i k(x - x_i) g(t - t_i|\sigma), \\ K_y = \sum_i p_i k(y - y_i) g(t - t_i|\sigma) \\
    \left\lgroup
        \begin{array}{l}
            K_x \\
            K_y
        \end{array}\right\rgroup = 
    \left\lgroup
        \begin{array}{l}
            \sum_i p_i k(x - x_i) g(t - t_i|\sigma) \\
            \sum_i p_i k(y - y_i) g(t - t_i|\sigma)
        \end{array}\right\rgroup,
\end{gather}
where $\sigma$ controls the temporal smoothing scale of the Gaussian kernel $g$: $g(a|\sigma) = exp(- (a / \sigma)^2)$. This projection strategically decouples spatial dimensions while maintaining fine temporal precision (\textit{e.g.}, 10$\mu s$) through continuous Gaussian temporal encoding, thereby enabling high-resolution observation of transient processes.

Our framework collaboratively combines both projection paradigms as spatio-temporal event representation: 1) Event Voxel captures spatially dominant features, and 2) Event Kymograph preserves temporally sensitive patterns.

\subsection{Spatio-temporal Dual Cost Volumes}
\textbf{Spatial Correlation}. Building upon established architectures in optical flow estimation \cite{gehrig2021raft, teed2020raft}, EMoTive employs two same 2D convolution networks as space encoder to extract latent spatial representation $f_{hw}:\mathbb{R}^{D_s \times H_D \times W_D}$ and context encoder to obtain context representation $f_{c}$ from the Event Voxel. 
% Our framework establishes all-pairs spatial cost volumes through pyramidal feature matching. 
And given latent features $f^p_{hw}$, $f^t_{hw}$ from temporally adjacent voxel projections, we generate an $N$-level cost pyramid via multi-resolution inner-product:
% \begin{align}
%    &C^m_{hw}(i, j, k, l) \notag\\
%    &= \frac{1}{2^{2m}} \sum_c \sum^{2^m}_{a} \sum^{2^m}_{b}  f^p_{hw}(c, i, j) \cdot f^t_{hw}(c, k + a, l + b) \notag \\
%     &= \sum_c f^p_{hw}(c, i, j) \cdot \frac{1}{2^{2m}} \sum^{2^m}_{a} \sum^{2^m}_{b} f^t_{hw}(c, k + a, l + b) \notag \\
%     &=  \sum_c f^p_{hw}(c, i, j) \cdot AP_{2d}(f^t_{hw}(c, k, l), 2^m).
% \end{align}
\begin{align}
   C^m_{hw} =  \frac{1}{D_s} \sum_{c=1}^{D_s} f^p_{hw}(c, i, j) \cdot \mathcal{A}_{2d}^{2m}(f^t_{hw}(c, k, l)),
\end{align}
where $\mathcal{A}_{2d}^{2^m}$ denotes 2D average pooling with kernel size $2^m$ for coarse-to-fine matching, $m \in \{1, \dots, N\}$ indexes the pyramid level, and we adopt a 2-level cost pyramid.

\textbf{Temporal Correlation}. For temporal feature extraction, we apply a 1D convolution encoder operating on the Event Kymograph $(K_x, K_y)$. The projection will firstly be evenly divided into subblocks based on the $N_a$ temporal anchors to form the sequence $K_{x|y,s} = \{K_{x|y,s}(1), \dots, K_{x|y,s}(N_a)\}$. The encoder operates on each subblock and outputs two complementary temporal feature tensors: $f_{ht}: \mathbb{R}^{N_a \times D_t \times H_D}$ and $f_{wt}: \mathbb{R}^{N_a \times D_t \times W_D}$, preserving axis-specific motion pattern. To further capture motion dynamics, we formulate a spatial-decoupled temporal cost pyramid through cross-subblock correlation along orthogonal spatio-temporal projections:
% Similarly, we construct a cost pyramid from  temporal features $f_{ht}, f_{wt}$ to guide subsequent cue extraction. In specific, for the temporal features in the $y-t$ diretion $f_{ht}$, the cost volumes between the first temporal subblock $f_{ht}(1)$ and subsequent temporal subblock $f_{ht}(n), n \in \{2, \dots, N_a\}$ are computed as follows:
\begin{align}
    \left\lgroup
        \begin{array}{l}
            C^m_{wt} \\
            C^m_{ht}
        \end{array}
    \right\rgroup= \left\lgroup
        \begin{array}{l}
            \frac{1}{D_t} \sum_{c} f_{wt}(1, c, i) \cdot \mathcal{A}_{1d}^{2^m}(f_{wt}(n, c, j)) \\
            \frac{1}{D_t} \sum_{c} f_{ht}(1, c, k) \cdot \mathcal{A}_{1d}^{2^m}(f_{ht}(n, c, l))
        \end{array},
    \right\rgroup
    % C^m_{wt}(n, i, j) =  \frac{1}{D_t} \sum_{c=1}^{D_t} f_{wt}(1, c, i) \cdot \mathcal{A}_{1d}(f_{wt}(n, c, j), 2^m),\\
    % C^m_{ht}(n, i, j) =  \frac{1}{D_t} \sum_{c=1}^{D_t} f_{ht}(1, c, i) \cdot \mathcal{A}_{1d}(f_{ht}(n, c, j), 2^m),
\end{align}
where, $\mathcal{A}_{1d}^{2^m}$ implements 1D temporal pooling, and $n \in \{2, \dots, N_a\}$ indexes temporal subblocks.

For subsequent efficient correlation sampling, the temporal correlation on the joint space axis is then synthesized through tensor product fusion: 
\begin{align}
    C^m_{t}(n, i, k, j, l) \doteq C^m_{ht}(n, i, j) \otimes C^m_{wt}(n, k, l),
\end{align}
where $\otimes$ denotes outer product along spacial dimensions.

\begin{figure}
    \centering
    \includegraphics[width=\linewidth]{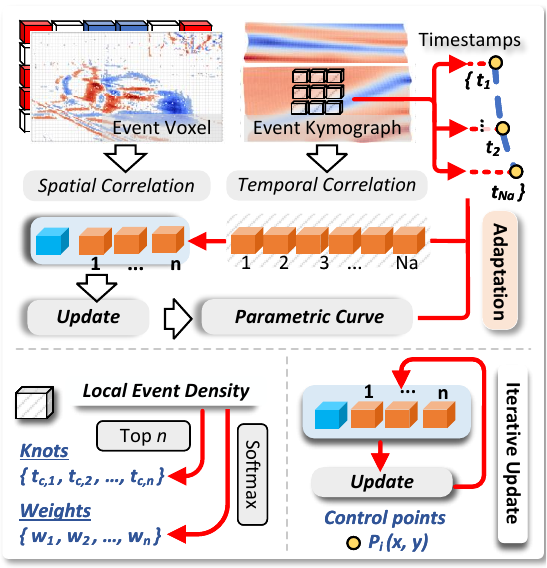}
    \caption{Event-guided trajectory formation process. Local event density is introduced to adaptively adjust the knots and weights while fusing spatio-temporal features for control points update.}
    \label{fig:pipeline}
\end{figure}

\subsection{Spatio-Temporal Trajectory}
% To model the heterogeneous spatio-temporal transformation, we refer to the designs in \cite{gehrig2024dense, tulyakov2022time} and use a parameterized curve to represent the spatio-temporal trajectory $\mathcal{T}$, updating the curve parameters via temporal querying. Unlike previous methods, to improve motion expression abilitys, we use a non-uniform rational B-spline curve to describe the spatio-temporal evolution process, replacing the traditional uniform or globally controlled curve description. The non-uniformity is modeled using event observations.
To model heterogeneous motion patterns from event stream, we propose a non-uniform rational B-spline (NURBS) based spatio-temporal trajectory representation. This parametric formulation extends prior uniform B-spline approaches \cite{gehrig2024dense, tulyakov2022time} by incorporating event-driven non-uniformity through adaptive knot vectors and weights, addressing limitations in the motion expressiveness of globally controlled curves. The trajectory map $\mathcal{T}: \mathbb{R}^3 \mapsto \ \mathbb{R}^2$ is defined as:
% \subsubsection{Parameterized Trajectory} The parameterized trajectory in the form of a non-uniform rational B-spline is given by: 
\begin{align}
    \mathcal{T}(t, x, y) = \frac{\sum_i^n  N_{i, p}(t) w_i \mathbf{P}_i(x, y)}{\sum_i^n N_{i, p}(t) w_i},
\end{align}
where $N_{i, p}(t)$ is the $p$-th degree B-spline basis function function with non-uniform knot vector $\mathbf{T} = \{t_1, \dots, t_m\}$, $w_i \in \mathbb{R}^{+}$ represents the event-adaptive weight and $\mathbf{P}_i \in \mathbb{R}^2$ is the $i$-th control point from the set $\mathcal{P} = \{\mathbf{P}_1, \dots, \mathbf{P}_{n} \}$. 
% The initial control point $\mathbf{P}_1$ is anchored at $(0, 0)$ to ensure boundary consistency. 
% For a clamped NURBS curve of degree $p$ with $n$ control points, the knot vector contains $m=n+p+1$ elements, with only $n-p$ intermediate knots $t_c \in(t_{p+1} ,t_{n+1})$ being adjustable. The trajectory $\mathcal{T}$ models pixel displacement through temporal evolution, where density-aware adaptation adjusts both knot vector distribution $\mathbf{T}$ and weight $\{w_i\}$ values, while control points are updated via spatio-temporal dual cost volumes.

% The spline basis function is expressed using the Cox-de Boor recursive formula: 

% \begin{align}
%     N_{i,0}(t) &= \left\{
%         \begin{array}{ll}
%             1 & \mbox{if } t_{i} \leq t < t_{i+1} \\
%             0 & \mbox{otherwise}
%     \end{array},
%     \right. \\
%     N_{i,p}(t) &= \frac{t - t_{i}}{t_{i+p} - t_{i}} N_{i,p-1}(t) \ + \notag\\ 
%     &\frac{t_{i+p+1} - t}{t_{i+p+1} - t_{i+1}} N_{i+1,p-1}(t),
% \end{align}

% where $t_i$ represents the node vector of the spline function, which corresponds to a particular node in the trajectory. Here, we uniformly refer to these as the node vectors. 

\subsubsection{Density-aware Adaptation}
The expressiveness of NURBS curves is primarily governed by knot vector distribution and control point weighting. Higher knot density, coupled with elevated weights, enhances the curve's capacity to capture rapid motion patterns. Building on the motion sensitivity of event \cite{gallego2020event}, we formulate an event-density guided adaptation mechanism.
% to adjust the knot vetor distribution and weight values.
The adaptation pipeline comprises three phases:

(1) Spatio-temporal density estimation: Compute the distribution tensor $E_s \in \mathbb{R}^{T\times H \times W}$ from Event Kymograph: $E_s(i_k, x, y) = K_{s, x}(i_k, x) \odot K_{s, y}(i_k, y)$,
% \begin{gather}
%     E_s(i_k, x, y) = K_{s, x}(i_k, x) \odot K_{s, y}(i_k, y)
% \end{gather}
where $i_k \in [1, N_a]$ indexes temporal blocks. A 3D average pooling operator $\mathcal{A}_{3d}$ generates the density $D_s = \mathcal{A}_{3d}(E_s)$.

(2) Key parameter selection: Extract top-$n$ temporal indices $\{i_{t,j}\}_{j=1}^n$ from  $D_s$ along the temporal axis, where $n$ matches the control point count. Then convert these to normalized timestamps: $t_{l,j} = \frac{i_{t,j}}{N_a}, \forall j \in {1,\dots,n}.$
% \begin{align}
%     t_{l,j} = \frac{i_{t,j}}{N_a}, \quad \forall j \in {1,\dots,n}
% \end{align}
% \begin{align}
%     t_{l,j} = i_{t,j} / N_a, \quad \forall j \in {1,\dots,n}
% \end{align}

(3) Curve parameter update: For clamped NURBS constraints, compute adjustable knots $t_{c, i}$ via sliding-window averaging and weight values from density normalization:
\begin{gather}
t_{c,i} = \frac{1}{p}\sum_{k=0}^{p-1} t_{l,i+k}, \quad i \in {1,\dots,n-p+1}, \\
\{w_j\}_{j=1}^n = \sigma\left({D_s(i_{t,j})}_{j=1}^n\right),
\end{gather}
where $\sigma(\cdot)$ denotes softmax normalization.

\subsubsection{Control Points}
The trajectory control points are dynamically updated through an event-guided spatiotemporal feature fusion process, as illustrated in \cref{fig:pipeline}. Our update mechanism operates in three phases:

\textbf{Spatio-temporal Query:} For each pixel $\mathbf{p} = (x, y)$ at  at reference time $t=0$, we sample its warped position $\mathbf{p}_t = \mathbf{p} + \mathcal{T}(t, x, y)$ along the NURBS trajectory. At each queried timestamp $\{t_{l,j}\}_{j=1}^n$ from density adaptation, we extract neighborhood correlation within radius $r$ from the cost pyramid.
% \begin{align}
% \mathcal{N}(\mathbf{p}_t) = \{\mathbf{p}_t + \delta\mathbf{p} \ | \ \delta\mathbf{p} \in \{-r,\ldots,r\}^2\}
% \end{align}
% where bilinear interpolation handles subpixel coordinates. 
This constructs 4D temporal/spatial cost volume $C \in \mathbb{R}^{H \times W \times n \times (2r+1)^2}$, capturing motion patterns. 

\textbf{Feature Fusion:} We combine three information streams: (1) Temporal cost volumes at adaptively sampled $\{t_{l,j}\}$; (2) Spatial cost volume from end-time displacement $\mathcal{T}(1, x, y)$;  (3) Context representation $f_c$ from spatial feature encoding. These are concatenated into a spatio-temporal feature $\mathcal{F} \in \mathbb{R}^{H \times W \times d}$ through convolutions.

\textbf{Iterative Refinement:} Initializing $\mathcal{P}^{(0)} = \mathbf{0}$, we iteratively update control points via feature $\mathcal{F}$: $ \mathcal{P}^{(j+1)} = \mathcal{P}^{(j)} + \text{GRU}\left(\mathcal{F}, \mathcal{P}^{(j)}\right)$, GRU is gated recurrent unit.

\subsection{Scene Flow Estimation}  
Building upon the depth-normalized scene flow $\hat{S}_f$ \cite{yang2020upgrading}, we decompose it into two complementary components: optical flow $\mathcal{O}$ and  motion in depth $\mathcal{M}$ :
\begin{align}
    \hat{S}_f = K \cdot \frac{S_f}{Z} = ((\mathcal{M} - 1) \cdot \mathbf{u} + \mathcal{M} \cdot \mathcal{O}),
\end{align}
where $K$ represents the camera intrinsic matrix, $Z$ denotes the relative scene depth, $\mathbf{u}$ indicates the homogeneous image coordinates, and $S_f$ corresponds to the scene flow. The optical flow component at time $\tau$ is directly computable through trajectory queries: $\mathcal{O}(x, y) = \mathcal{T}(\tau, x, y)$. 

The motion in depth component $\mathcal{M}$ can be derived from the temporal gradient of the trajectory $\mathcal{T}$. Assuming a non-rotating rigid body under perspective projection with constant velocity in world coordinates, we establish the depth motion relationship:
\begin{align}
    \frac{Z_1}{Z_0} = \mathcal{M} = \frac{v_0 \Delta t + \Delta x}{v_1  \Delta t + \Delta x},\ \Delta t = t_1 - t_0,
    % \left
    %     \begin{array}{c}
    %         \Delta t = t_1 - t_0   \\
    %         \Delta x = x_1 - x_0
    %     \end{array}\right),
    \label{eq:mid}
\end{align}
where $\Delta x$ and $(v_0, v_1)$ represent the displacement and velocities of the object along the x-axis between times $t_0$ and $t_1$, respectively (complete derivation provided in supplementary material). 

By estimating the trajectory gradient $\mathcal{T}'(t)$ \cite{pieglnurbs} to obtain instantaneous velocity $v$, we compute $\mathcal{M}$ via \cref{eq:mid}. Enforcing physical motion consistency across temporal observations yields the multi-view relationship: $\mathcal{M}_2 = \frac{t_2}{t_1}(\mathcal{M}_1 - 1) + 1$.
% \begin{align}
%     \mathcal{M}_2 = \frac{t_2}{t_1}(\mathcal{M}_1 - 1) + 1.
%     \label{eq:time}
% \end{align}
This leads to our temporal stability constraint through viewpoint aggregation:
\begin{gather}
    \mathcal{M}_{k} = \frac{1}{k} \sum_{i}\frac{t_k}{t_i}(\mathcal{M}_{i} - 1) + 1.
\end{gather}

\begin{figure*}[htbp]
    \centering
    \includegraphics[width=.95\linewidth]{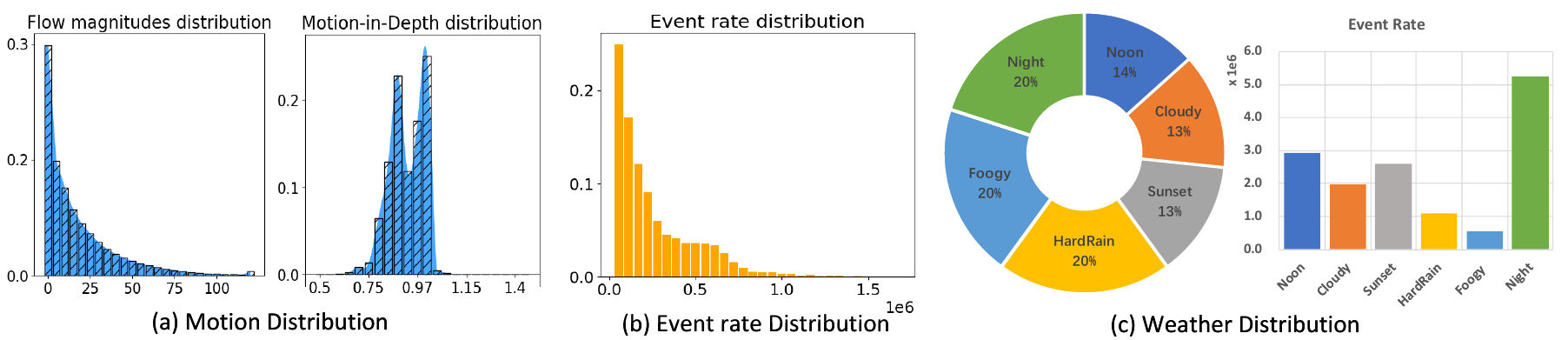}
    \caption{Characteristics of the data distribution in the CarlaEvent3D dataset}
    \label{fig:datasets_stat}
\end{figure*}

\begin{figure}[htbp]
    \centering
    \includegraphics[width=.97\linewidth]{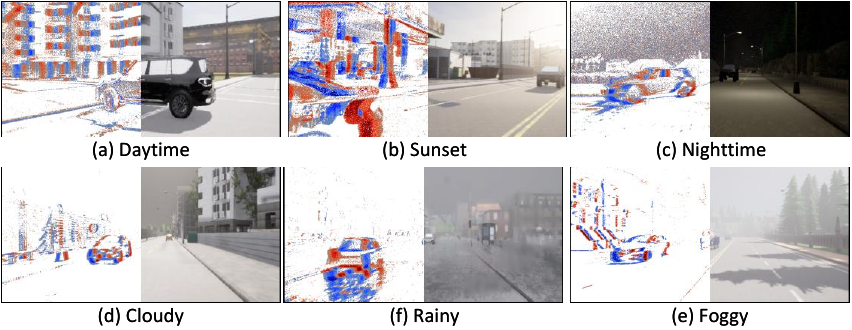}
    \caption{Event/image from CarlaEvent3D under multiple scenes.}
    \label{fig:samples}
\end{figure}

\subsection{Supervision Framework}
The proposed method employs a multi-task learning objective combining optical flow supervision, motion in depth supervision and temporal gradient regularization. The composite loss function is formulated as:
\begin{align}
L = L_{\text{flow}} + L_{\text{depth}} + \lambda L_{\text{t}}, \quad \lambda = 10^{-7}.
\end{align}

\textbf{Optical Flow Supervision}. We implement an exponentially weighted L1 loss across iterative refinements:
\begin{align}
L_{\text{flow}} = \sum_{k=1}^N \gamma^{N-k} \left( | \mathcal{O}_x^{(k)} |_1 + | \mathcal{O}_y^{(k)} |_1 \right),
\end{align}

\textbf{Motion in Depth Supervision}. Following similar principles, the motion in depth loss can be obtained as:
\begin{align}
L_{\text{depth}} = \sum_{k=1}^N \gamma^{N-k} | \mathcal{M}^{(k)} |_1.
\end{align}

\textbf{Temporal Gradient Regularization}. To prevent high-order trajectory distortions (\textit{e.g.}, spiral artifacts in linear motion), we impose a first-order smoothness constraint on temporal derivatives of trajectory:
\begin{align}
L_t = \sum_{i=1}^{T-1} | \mathcal{T}'(t_{i+1}) - \mathcal{T}'(t_i) |_1.
\end{align}
This regularization term enforces smooth temporal evolution of trajectory, effectively suppressing non-physical transformation while preserving motion consistency.

\section{Experiment}
\label{sec:exp}
\subsection{Experimental Setup}

\textbf{Dataset}. Current event-based 3D motion datasets predominantly derive ground truth through LiDAR in point cloud coordinates \cite{wan2023rpeflow, zhou2024bring}, resulting in sparse and inconsistent signals. These datasets exhibit several critical limitations, including a lack of dynamic object annotations and limited environmental diversity (daytime only). The DSEC benchmark \cite{gehrig2021dsec} provides improved event-aligned optical flow in corrected camera coordinates, supporting day-night scenarios. However, its filtering of moving objects restricts applicability to motion estimation. In this paper, we develop CarlaEvent3D dataset which provides various driving scenes and ground truth 3D motion annotations with Carla simulator. For event generation, we utilize this simulator to generate high-frame-rate video, then simulate events using the DVS Voltemter \cite{lin2022dvs} algorithm. The dataset contains 22,125 synchronized event-label tuples (resolution: 320×960) across 6 environmental conditions (\cref{fig:datasets_stat}), partitioned into: 1.Training: 45 sequences (13,275 samples); 2.Validation: 15 sequences (4,425 samples); 3.Testing: 15 sequences (4,425 samples). The collected objects guarantee a wide range of motion distribution, with velocity over 100 pixels / 100 ms (highway speeds). Furthermore, the distribution of events in various environments shows significant differences, from 0.5 - 5.0 Mev/s (foggy \textit{v.s.} night).

\begin{table*}
\centering
\small
\begin{tabular}{l|ccc|ccc|cc} 
\hline
                % & \multicolumn{3}{c|}{dense}                 & \multicolumn{3}{c|}{sparse}                &        &        \\ 
% \cline{2-7}
                & \multicolumn{2}{c}{flow@D} & mid@D & \multicolumn{2}{c}{flow@S} & mid@S &        &        \\ 
\hline
Model           & EPE$_\downarrow$      & f1$_\downarrow$        & log-mid$_\downarrow$       & EPE$_\downarrow$       & f1$_\downarrow$         & log-mid$_\downarrow$         & params.(M) & time(s)   \\ 
\hline
ERaft \cite{gehrig2021raft}          & 2.781 & 24.604           &    -            & 3.067  & 27.326          &     -           & 5.04   & 0.049 \\
Expansion \cite{yang2020upgrading}       & 7.821 & 57.653           & 171.237         & 8.266  & 61.819          & 173.781         & 12.13 & 0.300  \\
TPCV \cite{ling2023learning}           & 4.510 & 42.230           & 285.253         & 5.298  & 47.335          & 335.034         & 10.33 & 0.085  \\
ScaleFlow \cite{leng2023scaleflow}      & 4.518 & 42.885           & 268.050         & 5.288  & 47.956          & 321.067         & 10.70 & 0.090  \\
Scale++ \cite{ling2024scaleflow++}    & 5.242 & 40.081           & 260.165         & 6.044  & 46.391          & 298.588         & 42.96 & 0.119  \\
ETTCM$^1$ \cite{nunes2023time}          &   -   &    -             &   -             & 27.596 & 96.538          & 518.365         &   -    &   $10^{-6}/ev$  \\
\rowcolor{mygray}
\textbf{EMoTive(UniForm)} & 2.669 & 24.607           & 122.023         & 3.060  & 27.367     & 147.991        & 5.67  & 0.040  \\
\rowcolor{mygray}
\textbf{EMoTive}          & \textbf{2.547} & \textbf{22.866}           & \textbf{113.593}         & \textbf{2.850}  & \textbf{24.659}          & \textbf{138.402}         & 5.61  & 0.040  \\
\hline
% \small
% \multicolumn{9}{c}{\scriptsize{\makecell[l]{\textsuperscript{1}: It should be noted that ETTCM only outputs motion estimation results on each event, so it is unable to verify its performance in dense tags, and its number of parameters and time are proportional to the amount of events. The estimated time of each\\ event is $1 us$,. If the camera outputs $10 ^ {- 6} $events per second, its reasoning time will reach 1s, this comparison is not fair.}}}
\end{tabular}
\vspace{-.3cm}
\normalsize
\caption{The validation results on the CarlaEvent3D dataset. @D and @S represent dense labels and event-masked sparse labels.}
\label{table:CarlaEvent3d}

\end{table*}

% \begin{figure}
%     \centering
%     \begin{subfigure}[b]{\linewidth}
%          \centering
%          \includegraphics[width=1.\linewidth]{fig/weather_test.pdf}
%          \caption{Evaluation of Motion estimation performance in different environments. \textbf{\textit{N}}:Nightime, \textbf{\textit{D}}:Daytime, \textbf{\textit{S}}:Sunset, \textbf{\textit{C}}:Cloudy, \textbf{\textit{F}}:Foggy, \textbf{\textit{R}}:Rainy.}
%          \label{fig:scenes}
%      \end{subfigure}
%      \begin{subfigure}[b]{.75\linewidth}
%          \centering
%          \includegraphics[width=1.\linewidth]{fig/speed_test.pdf}
%          \caption{Speed tracking capability curve}
%          \label{fig:speeds}
%      \end{subfigure}
%     \caption{Verification of Motion estimation ability}
% \end{figure}

\textbf{Experimental Setup}.
We employ the AdamW optimizer with gradient clipping and the OneCycle learning rate policy over 60,000 iterations to ensure stable convergence. For the Event Kymograph projection kernel, we set the smoothing scale $\sigma=10$ to balance event density and temporal resolution. And the temporal subblock count $N_a$ is set to 6 for efficient feature aggregation. The hyperparameters of the trajectory will be defined through the ablation study.

The scene flow validation is partitioned into optical flow and motion in depth in CarlaEvent3D. For optical flow estimation, we adopt Average End-Point Error (EPE) and Fl-outlier ratio (Fl) \cite{gehrig2021raft} metrics, while motion in depth accuracy is quantified through logarithmic error (log-mid) \cite{yang2020upgrading}.
% where lower values correspond to improved estimation accuracy.  
Furthermore, we validate methods on the DSEC benchmark for scene flow estimation and employ 3D End-Point Error ($\text{EPE}_{\text{3D}}$) and Accuracy-at-5cm ($\text{ACC}_{0.05}$ metrics\cite{wan2023rpeflow}.

\subsection{Validation on the CarlaEvent3D Dataset}
In this section, we conducted comprehensive motion estimation benchmarking  on the CarlaEvent3D dataset, with comparisons detailed in \cref{table:CarlaEvent3d}. To ensure fair comparison under event-driven paradigms, we re-implement four image-based methods – Expansion \cite{yang2020upgrading}, TPCV \cite{yang2020upgrading}, ScaleFlow \cite{ling2022scale}, and Scale++ \cite{ling2024scaleflow++} – using Event Voxel as input with consistent training protocols.

The results reveal that the methods depending on spatial correlation exhibit significant performance degradation (5.274 - 1.963 EPE increase) due to event data's inherent low spatial redundancy,
% which fundamentally limits their spatial transformation reasoning capability.  
On the other hand, the ETTCM \cite{nunes2023time} model, which estimates motion per event, is limited by the asynchronous nature of the algorithm and struggles to achieve ideal performance in complex motion scenarios. Our EMoTive framework addresses these limitations through fine temporal observation and event-guided spatio-temporal trajectory modeling, outperforms spatial-dependent methods by 43.5\% in optical flow EPE, and 33.6\% in motion in depth log-mid. Furthermore, the event-guided non-uniform trajectory design specifically contributes 4.6\% (optical flow) and 6.9\% (depth motion) accuracy gains compared to uniform B-spline parameterization.

In terms of parameter count and inference time, EMoTive with only 5.61M learnable parameters (47.6\% fewer than ScaleFlow's 10.7M baseline), achieves 40 ms inference speed over 100 ms event data on an NVIDIA A6000 GPU, representing 52.9\% speed improvement.

\begin{table}[tbp]
\centering
\small
\begin{tabular}{l|cc|c} 
\hline
                & \multicolumn{2}{c|}{flow} & mid  \\ 
\hline
Model           & EPE$_\downarrow$   & f1$_\downarrow$             & log-mid$_\downarrow$         \\ 
\hline
ERaft \cite{gehrig2021raft}      &  0.451      &    0.773       &       -          \\
Expansion \cite{yang2020upgrading}      & 1.590  & 6.808            & 179.044          \\
TPCV \cite{ling2023learning}           & 0.782  & 2.353            & 379.688          \\
ScaleFlow \cite{leng2023scaleflow}      & 1.047  & 4.733            & 367.939          \\
Scale++ \cite{ling2024scaleflow++}    & 1.106  & 6.464            & 345.220          \\
ETTCM \cite{ling2024scaleflow++}          & 16.917 & 84.318           & 433.3            \\ 
\rowcolor{mygray}
\textbf{EMoTive(UniForm)} & 0.497  & 1.083   & 156.164          \\
\rowcolor{mygray}
\textbf{EMoTive}          & \textbf{0.439}  & \textbf{0.742}   & \textbf{152.722}          \\
\hline
\end{tabular}

\caption{Validation of 3D motion estimation on the DSEC dataset}
\label{table:DSEC}

\end{table}
% \begin{table}
% \centering
% \setlength{\tabcolsep}{3pt}
% \begin{tabular}{l|c|cc} 
% \hline\hline
% Model     & Input               & EPE$_{3D\,\downarrow}$   & ACC$_{.05\,\uparrow}$   \\ 
% \hline
% CamLiFlow & RGB+PC         & 0.120 & 53.5\%  \\
% RPEFlow   & RGB+PC+Ev & 0.104 & 60.5\%  \\ 
% \hdashline
% Expansion & Ev               & 0.489 & 4.3\%   \\
% Scale++   & Ev               & 0.603 & 1.4\%   \\
% \rowcolor{mygray}
% EMoTive    & Ev               & 0.205 & 25.5\%  \\
% \hline\hline
% \end{tabular}
% \caption{在 DSEC 数据集上的场景流估计验证。 “PC” 指代雷达点云，“Ev” 指事件流。}
% \end{table}

\begin{figure}[htb]
    \centering
    \small
    \setlength{\tabcolsep}{3pt}
    \begin{tabular}{l|c|cc} 
    \hline
    Model     & Input               & EPE$_{3D\,\downarrow}$   & ACC$_{.05\,\uparrow}$   \\ 
    \hline
    CamLiFlow \cite{liu2022camliflow} & RGB+PC         & 0.120 & 53.5\%  \\
    RPEFlow \cite{wan2023rpeflow}  & RGB+PC+Ev & 0.104 & 60.5\%  \\ 
    \hdashline
    Expansion \cite{yang2020upgrading} & Ev               & 0.489 & 4.3\%   \\
    Scale++ \cite{ling2024scaleflow++}  & Ev               & 0.603 & 1.4\%   \\
    \rowcolor{mygray}
    EMoTive    & Ev               & 0.205 & 25.5\%  \\
    \hline
    \end{tabular}
    \captionof{table}{Quantitative evaluation of scene flow on DSEC datasets. "PC" refers to the radar point cloud, and "Ev" refers to the event.}
    \label{table:sf}
    
    \vspace{.2cm}
    
    \includegraphics[width=.95\linewidth]{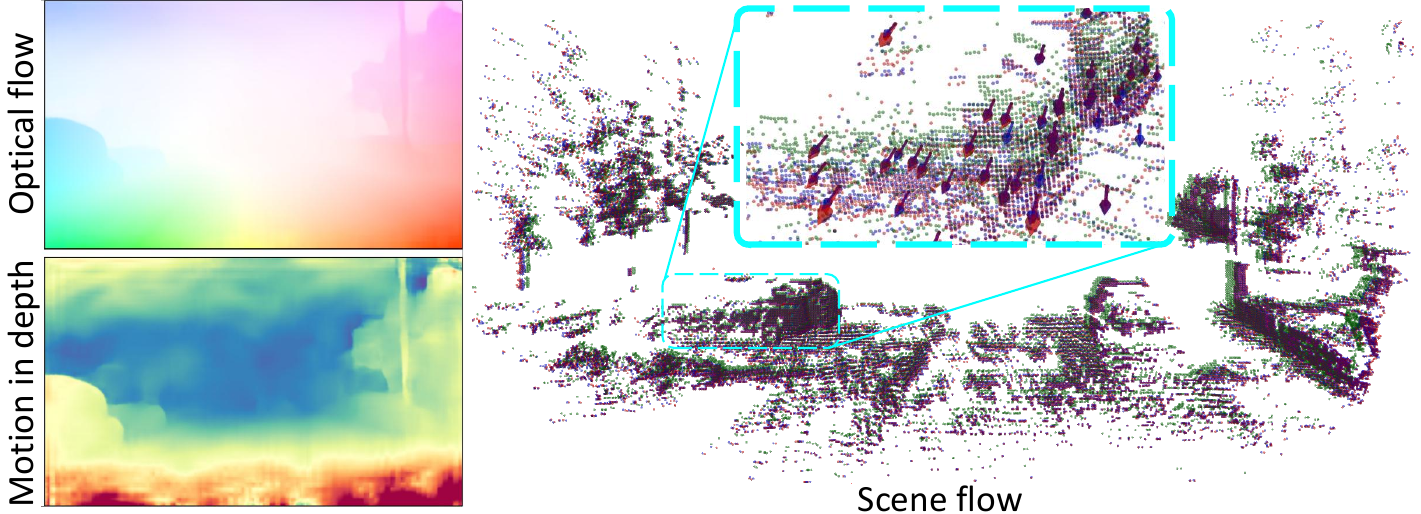}
    \captionof{figure}{Scene flow estimation. The green point cloud represents the initial position, the red is the position at the target time, and the red arrow denotes the real scene flow. The blue point cloud and arrows are the estimated position and corresponding scene flow.}
    \label{fig:sf}
    
\end{figure}

\subsection{Validation on the DSEC Dataset}

% \subsection{Real-world Validation on DSEC Benchmark}  
\label{sec:dsec_validation}  
To validate our method's practical efficacy, we conduct 3D motion and scene flow estimation on the DSEC dataset. Following the dataset split of \cite{wan2023rpeflow}, we train ERaft \cite{gehrig2021raft} from scratch due to its unavailable official implementation. All methods are evaluated under identical conditions.

\textbf{3D Motion Estimation }. As shown in \cref{table:DSEC}, EMoTive shows superior performance with 0.439px EPE (43.9\% improvement over TPCV), 0.742\% f1 ratio (1.611\% reduction), and 152.722 log-mid error (14.7\% better than Expansion). The spatial sparsity challenges methods like TPCV, whose performance degrades by 33.1\% in motion in depth due to its spatial reliance.  

\textbf{Scene Flow Analysis}. 
The scene flow evaluation in \cref{table:sf} reveals that EMoTive surpasses monocular event-based methods by 58\%, achieves $\text{EPE}_{\text{3D}}=0.205$cm and $\text{ACC}_{0.05}=25.5\%$. However, the multimodal fusion methods (CamliFlow/RPEFlow) maintain 41.5-49.3\% advantage due to privileged depth from Lidar \cite{wan2023rpeflow}. Furthermore, the scene flow projections in \cref{fig:sf} demonstrate EMoTive's capability to catch motion and structure in 3D.

\subsection{Ablation Study}
\label{sec:ablation}

\begin{figure}[tp]
    \centering
    \includegraphics[width=\linewidth]{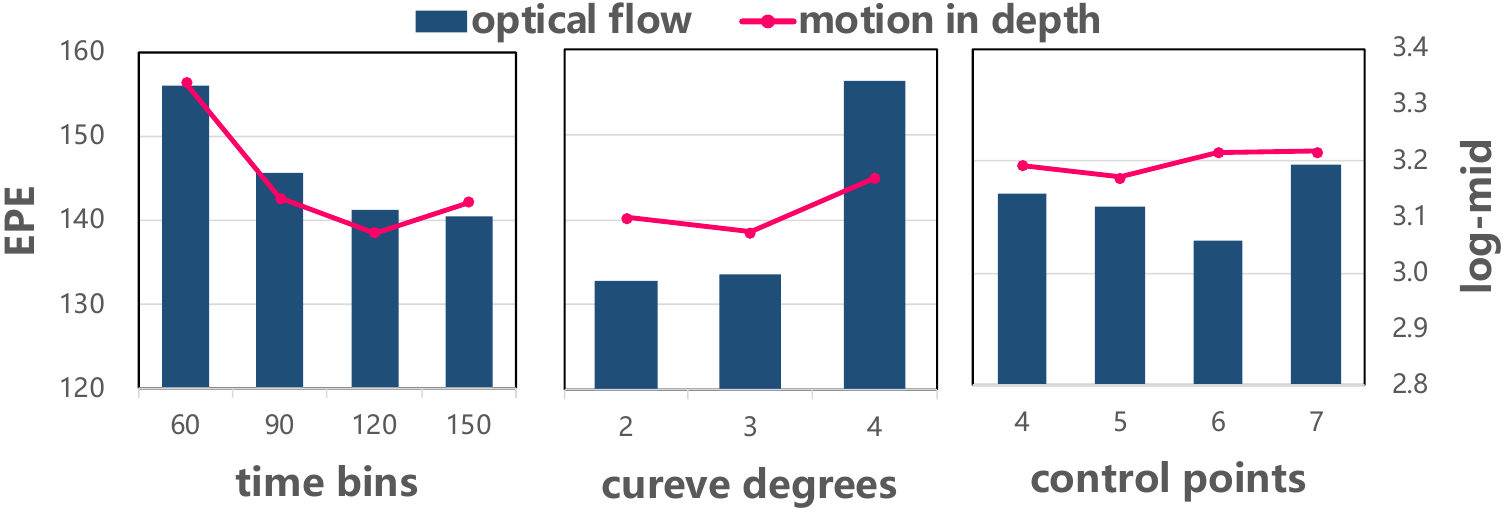}
    \caption{Impact of the trajectory parameters.}
    \label{fig:abl}
\end{figure}

% \begin{table}[htb]
% \centering
% \begin{tabular}{c|cc} 
% \hline\hline
%                       & flow  & mid      \\ 
% \hline
% Trajectory constraint & EPE   & log-mid  \\ 
% \hline
% \ding{56}               & 3.148 & 139.409  \\
% \ding{52}                  & 3.118 & 138.585  \\
% \hline\hline
% \end{tabular}
% \end{table}

% \begin{figure}[htb]
%     \centering
%     \small
%     \begin{tabular}{c|cc} 
%     \hline\hline
%                           & flow  & mid      \\ 
%     \hline
%     Traj $\mathcal{T}(t)$ constraint $L_t$ & EPE   & log-mid  \\ 
%     \hline
%     \ding{56}               & 3.148 & 139.409  \\
%     \ding{52}                  & 3.118 & 138.585  \\
%     \hline\hline
%     \end{tabular}
%     \vspace{-.3cm}
%     \captionof{table}{The impact of trajectory consistency constraints}
%     \label{table:loss}
    
%     \vspace{.2cm}
    
%     \includegraphics[width=.8\linewidth]{fig/traj_loss.pdf}
%     \vspace{-.2cm}
%     \captionof{figure}{Trajectory comparison under first-order neighborhood constraints}
%     \label{fig:loss}
% \end{figure}

\begin{figure}
    \begin{minipage}[h]{.24\textwidth}
        \centering
        \includegraphics[width=\textwidth]{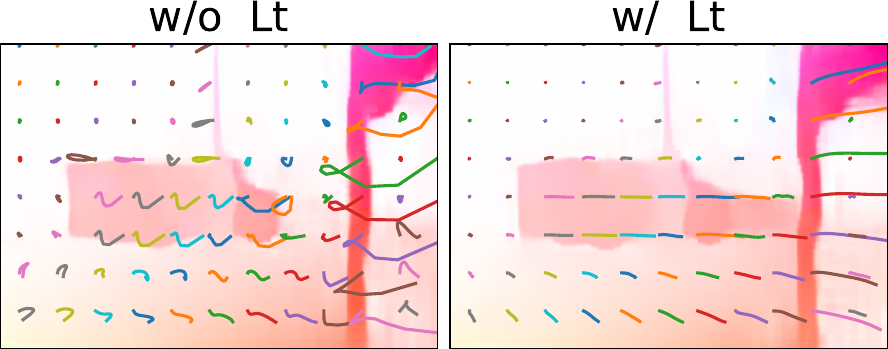}
    \end{minipage}
	\begin{minipage}[h]{.2\textwidth}
        \centering
        \small
        \setlength{\tabcolsep}{3pt}
        \begin{tabular}{c|cc} 
        \hline
                              & flow  & mid      \\ 
        \hline
       $L_t$ & EPE   & log-mid  \\ 
        \hline
        \ding{56}               & 3.148 & 139.409  \\
        \ding{52}                  & 3.118 & 138.585  \\
        \hline
        \end{tabular}
    \end{minipage}

    \captionof{table}{Impact of trajectory regularization on EMoTive.}
    \label{fig:loss}
\end{figure}

\begin{figure}
    \centering
    \includegraphics[width=.98\linewidth]{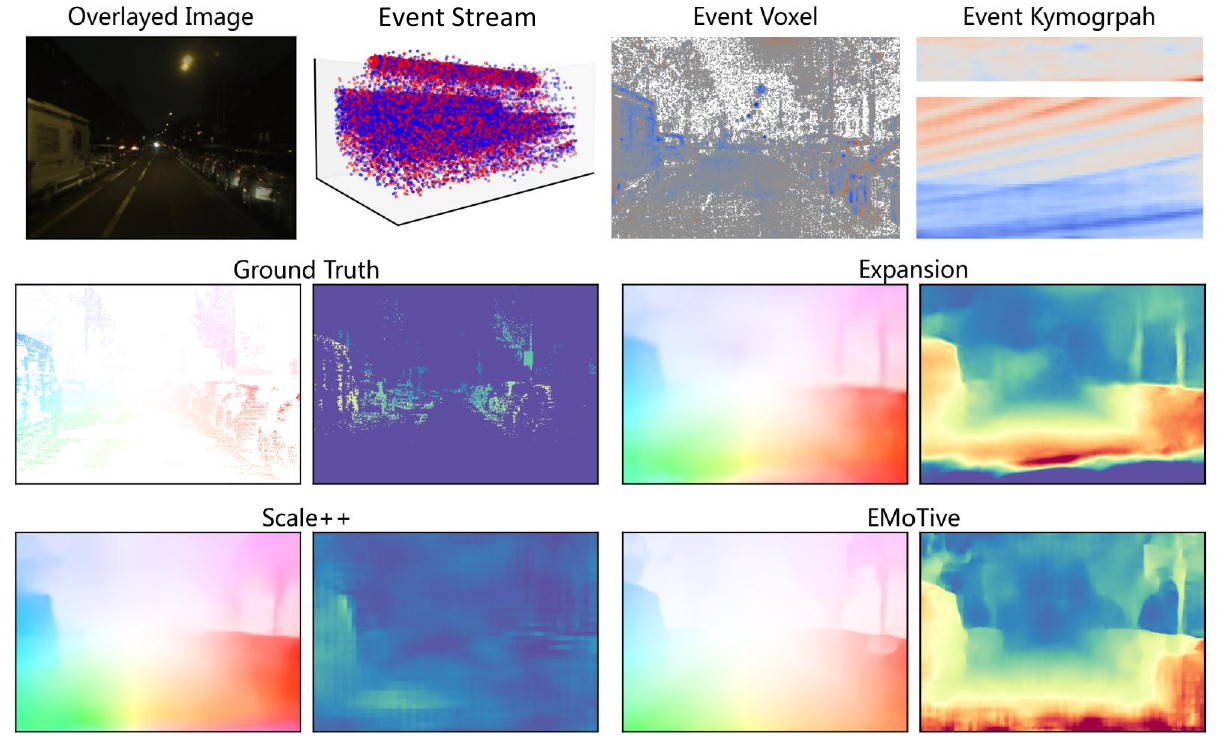}
    \caption{Visual comparison of motion estimation results (DSEC). The left is optical flow and the right is motion in depth estimation.}
    \label{fig:DSEC_night}
\end{figure}

To systematically validate EMoTive's architectural design, we conduct several ablation experiments on CarlaEvent3D's challenging scenarios (\textit{e.g.}, rainy days, night scenes, \textit{etc.}). Four critical components are analyzed: temporal resolution, trajectory degrees, control point number, and constraint loss ($L_t$). All variants are trained from scratch with identical hyper-parameters.

\textbf{Temporal Resolution Analysis}.
The temporal resolution in the Event Kymograph affects motion characterization. As shown in \cref{fig:abl}, we observe that the performance keeps improving with finer resolution as it better captures motion. However, when beyond 120 bins, noise amplification in temporal gradients degrades accuracy. Therefore, the optimal time granularity is 120 bins / 100ms.

\textbf{Trajectory degree}.
The curve degree $p$ of event-guided trajectories governs motion representation fidelity. As shown in \cref{fig:abl}, we observe that 3rd-order is optimal, where higher orders introduce overfitting in optical flow while lower orders underfitting in motion in depth.

\textbf{Control Point}.
The control point count $N_c$ balances representational capacity and over-smoothing risks.  As quantified in \cref{fig:abl}, the optimal $N_c=5$ achieves 3.118px EPE (0.024px decrease from $N_c=4$) and 138.585 log-mid error (1.6\% improvement over $N_c=6$). 

\textbf{Trajectory Regularization}.
The temporal gradient consistency constraint suppresses high-order distortions in trajectory learning. Without this constraint, as visualized in \cref{fig:loss}, trajectories exhibit undesirable curling due to unbounded higher-order derivatives. Quantitative results in \cref{fig:loss} reveal subtle but critical improvements, 0.03 error decrease in EPE and 0.824 decrease in log-mid.

\begin{figure}
    \centering
    \includegraphics[width=.95\linewidth]{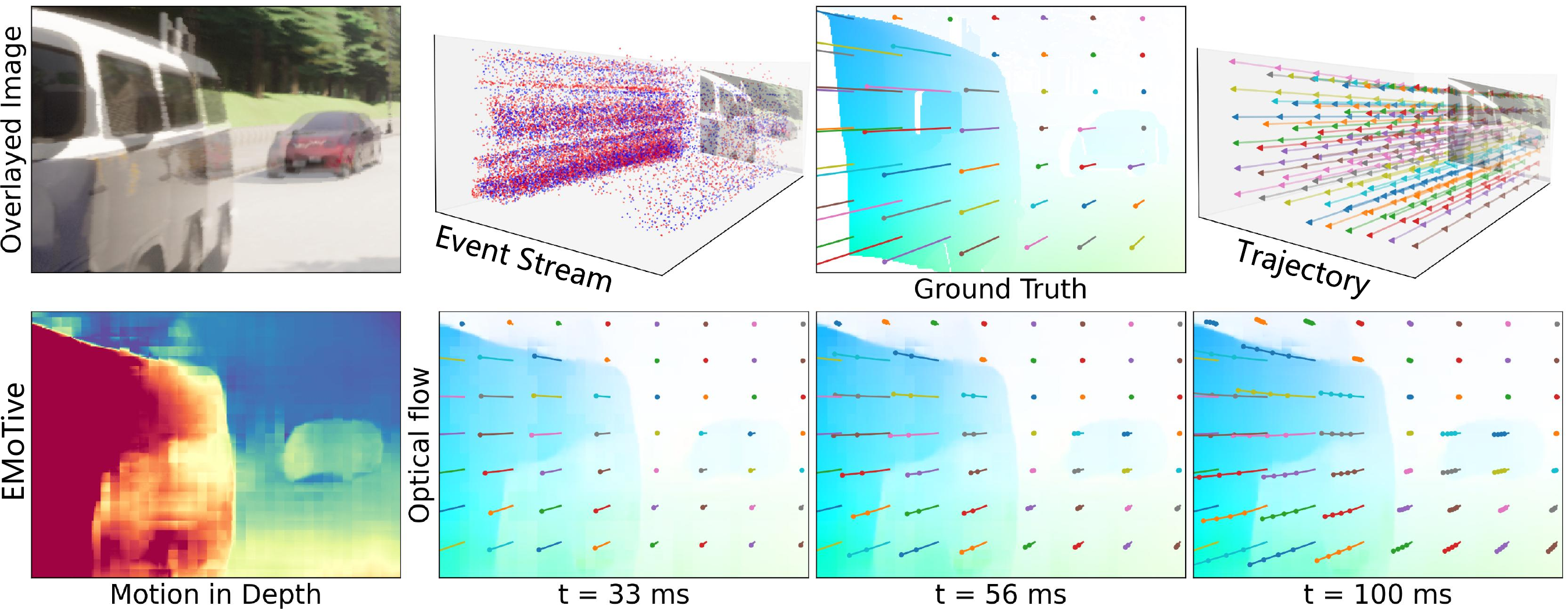}
    \caption{Visualization of trajectory estimated from EMoTive. Optical flow and motion in depth are derived from the trajectory.}
    \label{fig:traj}
\end{figure}

\subsection{Result Visualization}
To intuitively demonstrate the motion estimation performance, we visualized the results on the DSEC dataset in a nighttime scene, with comparative results in \cref{fig:DSEC_night}. Due to dark noise, there is severe degradation in the event projections, which leads to suboptimal motion estimation from the model only relying on spatial features. However, EMoTive maintains sharp motion boundaries, yielding better motion results. This robustness stems from event-guided dual spatio-temporal features fusion and parametric representation of trajectories, introducing motion knowledge on the spatio-temporal process. 

Furthermore, we visualized the trajectories estimated by EMoTive in \cref{fig:traj}. The trajectories not only provide motion in depth but also allow querying of the motion at any moment. Typical optical flow results assume that planar motion remains first-order invariant, only capturing linear motion. In contrast, the trajectory curves from EMoTive express high-order motion, better reflecting the heterogeneous spatio-temporal changes.

\section{Conclusion}
\label{sec:conclu}
This paper addresses the fundamental challenge of local spatio-temporal inconsistent motion in 3D motion estimation through a novel trajectory-centric framework. The proposed EMoTive framework establishes temporal continuity in motion representation by developing parametric curve trajectories guided by events. The technical foundation is built upon two pillars: 1) A fine-grained temporal projection, termed Event Kymograph, that leverages a continuous temporal projection kernel while decoupling spatial observations, and 2) A non-uniform rational curve parameterization framework that adaptively models trajectories under event guidance. The introduced temporal gradient regularization further ensures the physical plausibility of the learned trajectories without sacrificing representational capacity. Comprehensive evaluation on both synthetic and real-world benchmarks demonstrates superior performance in challenging scenarios. Future research will focus on combining this paradigm with other sensors and exploring its integration with predictive control systems.

{
    \small
    \bibliographystyle{ieeenat_fullname}
    \bibliography{main}
}

\clearpage
\setcounter{page}{1}
\maketitlesupplementary

\section{The Details of EMoTive Pipeline}
\label{sec:rationale}

Here, we provide more details about the EMoTive framework.

\subsection{Spatio-temporal Feature Encoder}
\textbf{Spatial Feature Encoding}. Building upon established architectures in optical flow estimation \cite{gehrig2021raft, teed2020raft},  EMoTive employs two same 2D convolutional networks to extract latent spatial representation $f_{hw}:\mathbb{R}^{D_s \times H_D \times W_D}$ and context representation $f_{c}$ from the Event Voxels. Specifically, we obtain the two Event Voxels from  adjacent time $V^p : \mathbb{R}^{B \times H \times W}, V^t : \mathbb{R}^{B \times H \times W}$. The bins $B$ is set to 7. Both voxels are processed by the spatial feature encoder to extract the spatial information at adjacent time, while the latter one is processed by the context feature encoder in the meantime for initializing motion information and mask features. The spatial feature encoder uses a 2D spatial convolutional layer with residual blocks to downsample the Event Voxel $V$, producing spatial features $f_{hw}:\mathbb{R}^{D_s \times H_D \times W_D}$. Here, $H_D = H / 8, W_D = W / 8$. The context feature encoder uses the same structure but outputs initialized motion information and mask features for upsampling via \textit{ReLU} and \textit{Tanh} activation functions.

\textbf{Temporal Feature Encoding}. For temporal feature extraction, we apply a 1D convolution encoder operating on the Event Kymographs $(K_x, K_y)$. The whole encoder framework is designed following the spatial encoder, except replacing the 2D convolution with 1D. The Event Kymographs will firstly be evenly divided into subblocks based on the $N_a$ temporal anchors to form the sequence $K_{x|y,s} = \{K_{x|y,s}(1), \dots, K_{x|y,s}(N_a)\}$. The encoder operates on each subblock and outputs $D_t$-dimensional temporal features. Meanwhile, progressive spatial downsampling is performed to align with the spatial feature. This encoder outputs two complementary temporal feature tensors: $f_{ht}: \mathbb{R}^{N_a \times D_t \times H_D}$ and $f_{wt}: \mathbb{R}^{N_a \times D_t \times W_D}$,  preserving axis-specific motion pattern.

\subsection{Spatio-Temporal Trajectory}

The parameterized trajectory in the form of a non-uniform rational B-spline is given by: 
\begin{align}
    \mathcal{T}(t, x, y) = \frac{\sum_i^n  N_{i, p}(t) w_i \mathbf{P}_i(x, y)}{\sum_i^n N_{i, p}(t) w_i},
\end{align}
where $N_{i, p}(t)$ is the $p$-th degree B-spline basis function function with non-uniform knot vector $\mathbf{T} = \{t_1, \dots, t_m\}$, $w_i \in \mathbb{R}^{+}$ represents the event-adaptive weight and $\mathbf{P}_i \in \mathbb{R}^2$ is the $i$-th control point from the set $\mathcal{P} = \{\mathbf{P}_1, \dots, \mathbf{P}_{n} \}$. 

The spline basis function in this paper adopts the Cox-de Boor recursive formula: 

\begin{align}
    N_{i,0}(t) &= \left\{
        \begin{array}{ll}
            1 & \mbox{if } t_{i} \leq t < t_{i+1} \\
            0 & \mbox{otherwise}
    \end{array},
    \right. \\
    N_{i,p}(t) &= \frac{t - t_{i}}{t_{i+p} - t_{i}} N_{i,p-1}(t) \ + \notag\\ 
    &\frac{t_{i+p+1} - t}{t_{i+p+1} - t_{i+1}} N_{i+1,p-1}(t),
\end{align}
where $t_i$ represents the node vector. For a clamped NURBS curve of degree $p$ with $n$ control points, the knot vector contains $m=n+p+1$ elements, with only $n-p$ intermediate knots $t_c \in(t_{p+1} ,t_{n+1})$ being adjustable. The trajectory $\mathcal{T}$ models pixel displacement through temporal evolution, where density-aware adaptation adjusts both knot vector distribution $\mathbf{T}$ and weight $\{w_i\}$ values, while control points are updated via spatio-temporal dual cost volumes.

\subsubsection{Spatio-temporal Query} 
For each pixel $\mathbf{p} = (x, y)$ at  at reference time $t=0$, we sample its warped position $\mathbf{p}_t = \mathbf{p} + \mathcal{T}(t, x, y)$ along the NURBS trajectory. At each queried timestamp $\{t_{l,j}\}_{j=1}^n$ from density adaptation, we extract neighborhood correlation within radius $r$ from the cost pyramid:
\begin{align}
    \mathcal{N}(\mathbf{p}_t) = \{\mathbf{p}_t + \delta\mathbf{p} \ | \ \delta\mathbf{p} \in \{-r,\ldots,r\}^2\},
\end{align}
where bilinear interpolation handles subpixel coordinates. The radius is set to 4 in this paper.

\subsubsection{Feature Fusion} 
We combine three information streams for control point update: (1) Temporal cost volumes at adaptively sampled $\{t_{l,j}\}$; (2) Spatial cost volume from end-time displacement $\mathcal{T}(1, x, y)$;  (3) Context representation $f_c$ from spatial feature encoding. These are concatenated into a spatio-temporal feature $\mathcal{F} \in \mathbb{R}^{H \times W \times d}$ through convolutions. The temporal cost volumes and spatial cost volume will be concated as the dual spatio-temporal cost volume, combined with parameters of control points to get the latent motion feature from motion encoder. The motion encoder is composed of dual branches  with 2D convolution. Then, the latent motion feature is concated with context representation $f_c$ to compose the final spatio-temporal feature. The number of control point iterative refinements is set to 6.

\subsection{Motion in Depth Estimation}
The motion in depth component $\mathcal{M}$ can be derived from the temporal gradient of the trajectory $\mathcal{T}$. Assuming a non-rotating rigid body under perspective projection with constant velocity in world coordinates, we establish the depth motion relationship:
\begin{align}
    \frac{Z_1}{Z_0} = \mathcal{M} = \frac{v_0 \Delta t + \Delta x}{v_1  \Delta t + \Delta x},\ \Delta t = t_1 - t_0,
\end{align}
where $\Delta x$ and $(v_0, v_1)$ represent the displacement and velocities of the object along the x-axis between times $t_0$ and $t_1$, respectively (complete derivation provided in supplementary material).

To elaborate further, consider a common scenario: a non-rotating rigid body using a pinhole camera model, the velocity of an object relative to the camera along the x-axis can be expressed as:
\begin{align}
    v = \frac{V_x - x V_z}{Z}.
    \label{eq:v}
\end{align}
Here, we normalize the pixel coordinates $x_{pix}$ and focal length $f$ such that $x = \frac{x_{pix}}{f}$. $Z$ represents the object's relative depth in the camera coordinate system, while $V_x$ and $V_z$ denote the 3D motion velocities of the object along the x-axis and z-axis, respectively. The term $v$ represents the instantaneous velocity along the x-axis in the image plane, with the same expression applying to the velocity projection along the y-axis, denoted as $u$. Assuming the object's 3D motion velocity remains constant in the real-world coordinate, we examine the projected velocity $v_0$ and $v_1$ at times $t_0$ and $t_1$, respectively. This leads to the following equation:
\begin{gather}
    v_0 - v_1 \frac{Z_1}{Z_0} = \frac{x_1 - x_0}{t_1 - t_0}(\frac{Z_1}{Z_0} - 1),
\end{gather}
where $x_0$ and $x_1$ represent the positions of the object along the x-axis at times $t_0$ and $t_1$, respectively. And the velocities along the z-axis, $V_z$ in \cref{eq:v} is replaced under the constant assumption:
$V_z = \frac{Z_1 - Z_0}{t_1 - t_0}$. By re-organizing the above equation, we can obtain the form of motion in depth as follows:
\begin{gather}
    \frac{Z_1}{Z_0} = \mathcal{M} = \frac{v_0 (t_1 - t_0) + (x_1 - x_0)}{v_1 (t_1 - t_0) + (x_1 - x_0)}.
\end{gather}
Given the starting observation position and time $x_0, t_0 $ set to 0, we can obtain:
\begin{align}
    \mathcal{M} = \frac{v_0 t_1 + x_1}{v_1 t_1 + x_1}
    \label{eq:mid}
\end{align}
Once the expression for the object's motion trajectory $\mathcal{T}$ is obtained, its time gradient $\mathcal{T}'(t)$ can be estimated, corresponding to the instantaneous velocity $(v, u)$. The gradient of the parameter \( t \) for the non-uniform rational B-spline (NURBS) curve used in this paper is given by the following:
\begin{gather}
    (v, u) = \mathcal{T}'(t) = \notag \\
    \frac{\sum_i N_{i, p}'(t) w_i \mathbf{P}_i(x, y) - \mathcal{T}(t)\sum_i N_{i, p}'(t) w_i}{\sum_i N_{i, p}(t) w_i }, \\
    N_{i, p}'(t) = \frac{p}{t_{i + p} - t_i} N_{i, p -1} - \frac{p}{t_{i + p + 1} - t_{i + 1}} N_{i +  1, p -1}.
    \label{eq:grad}
\end{gather}

Combined with \cref{eq:mid}, the estimation of motion in depth based on the trajectory is as follows:
\begin{align}
    \mathcal{M} = \frac{\mathcal{T}'(t_0) t_1 + \mathcal{T}(t_1)}{\mathcal{T}'(t_1) t_1 + \mathcal{T}(t_1)} = 1 + \frac{\mathcal{T}'(t_0) t_1 - \mathcal{T}'(t_1) t_1}{\mathcal{T}'(t_1) t_1 + \mathcal{T}(t_1)}
\end{align}

The final multi-view estimation algorithm of motion in depth is shown in \cref{algo:mid}.

\begin{algorithm}
    \renewcommand{\algorithmicrequire}{\textbf{INPUT:}}
    \renewcommand{\algorithmicensure}{\textbf{OUTPUT:}}
    \caption{Motion in Depth Multi-View Estimation Process}
    \label{algo:mid}
    \begin{algorithmic}[1]
        \REQUIRE Trajectory $\mathcal{T}(t)$, timestamps: $\{t_0, t_1, \dots, t_k\}$
        \ENSURE Motion-in-depth at \( t_k \): $\mathcal{M}_k$
        \STATE Retrieve trajectory values at each time point $\{\mathcal{T}(t_0), \mathcal{T}(t_1), \dots, \mathcal{T}(t_k)\}$
        \STATE Compute trajectory time gradients (instantaneous velocities) at each time point according to \cref{eq:grad} $\{\mathcal{T}'(t_0), \mathcal{T}'(t_1), \dots, \mathcal{T}'(t_k)\}$
        \STATE Multiply the initial instantaneous velocity by each timestamp to obtain the initial path estimate $\{\mathcal{T}'(t_0) t_1, \mathcal{T}'(t_0) t_2, \dots, \mathcal{T}'(t_0) t_k\}$
        \STATE Multiply the instantaneous velocity at each time point by the respective timestamp to get the endpoint path estimate $\{\mathcal{T}'(t_1) t_1, \mathcal{T}'(t_2) t_2, \dots, \mathcal{T}'(t_k) t_k\}$
        \STATE Following \cref{eq:mid}, combine the trajectory, initial path estimate, and endpoint path estimate to compute depth motion estimates at each time point $\{\mathcal{M}_1, \mathcal{M}_2, \dots, \mathcal{M}_k\}$
        \STATE According to the multi-view relationship, convert depth motion observations from different times to the same time point, yielding a series of depth motion estimation at time \( t_k \): $\{\mathcal{M}_{1, k}, \mathcal{M}_{2, k}, \dots, \mathcal{M}_{k, k}\}$
        \STATE Combine historical observations from different time points to stabilize the depth motion estimate at \( t_k \): $\mathcal{M}_{k} = \frac{1}{k} \sum_{i}{\mathcal{M}_{i, k}}$
    \end{algorithmic}
\end{algorithm} 

\subsection{Parameter Upsampling}
Since the spatiotemporal information for motion estimation comes from downsampled features, its output (including motion trajectory, optical flow, and depth motion) is at 1/8 of the original resolution. In this paper, the mask obtained from the context feature encoder is used to upsample the motion estimation output to full resolution. Specifically, the motion parameters at lower resolutions are first expanded using a $3 \times 3$ grid, and then a convex combination is performed based on the mask (which has been normalized per channel using the Softmax function) to upsample and obtain the motion estimation results at full resolution.

\begin{figure}[htbp]
    \centering
    \includegraphics[width=.95\linewidth]{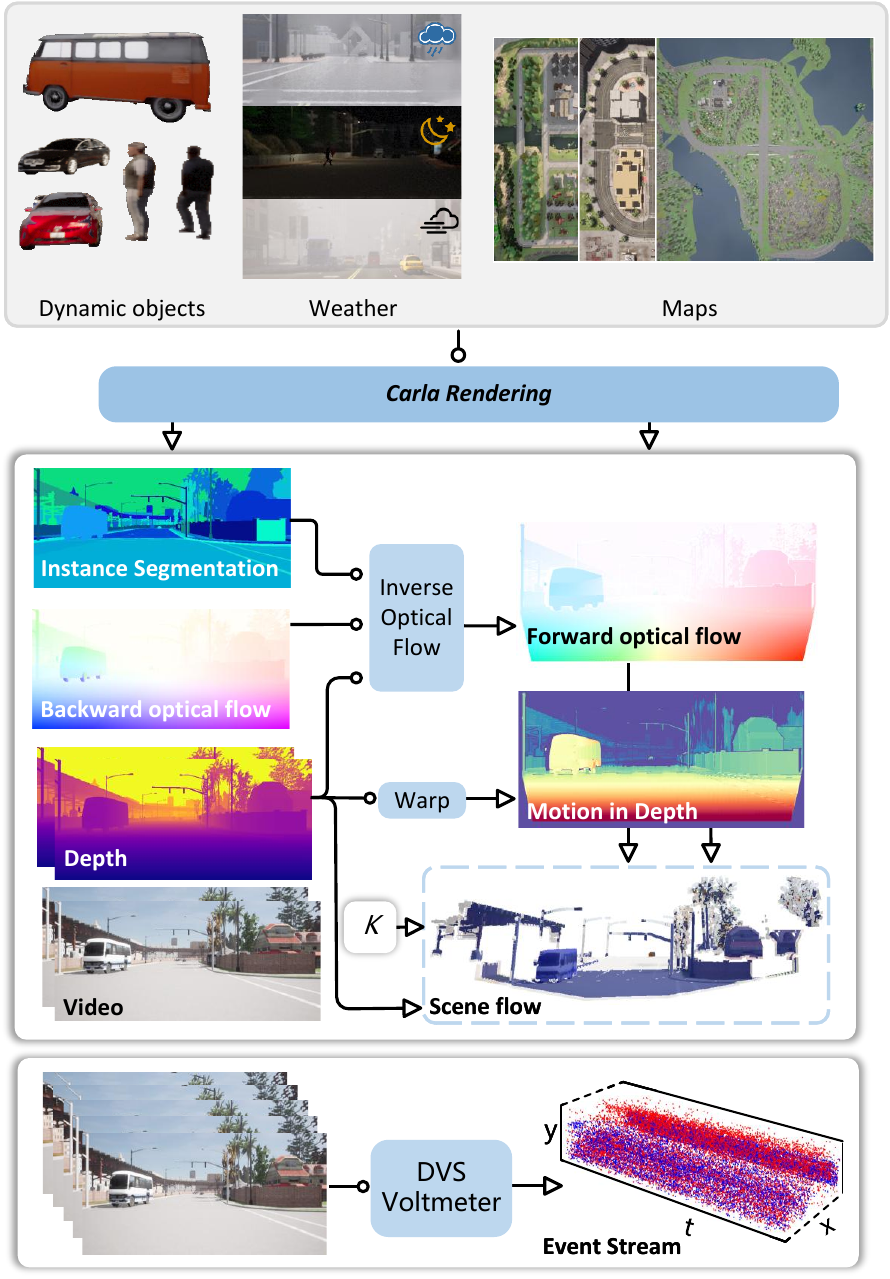}
    \caption{Data Simulation process based on Carla Simulator}
    \label{fig:simulation}
\end{figure}

\begin{table*}
\centering
\small
\setlength{\tabcolsep}{2.5pt}
\begin{tabular}{l|c|c|c|c|c|ccc|ccc} 
\hline\hline
\multirow{3}{*}{Datasets} & \multirow{3}{*}{year} & \multirow{3}{*}{\begin{tabular}[c]{@{}c@{}}Synthetic\\/Real\end{tabular}} & \multirow{3}{*}{\begin{tabular}[c]{@{}c@{}}Optical \\Flow\end{tabular}} & \multirow{3}{*}{Depth} & \multirow{3}{*}{\begin{tabular}[c]{@{}c@{}}Instance \\Segmentation\end{tabular}} & \multicolumn{6}{c}{Environment}                               \\ 
\cline{7-12}
                          &                       &                                                                           &                                                                         &                        &                                                                                  & \multicolumn{3}{c|}{Lighting}  & \multicolumn{3}{c}{Weather}  \\ 
\cline{7-12}
                          &                       &                                                                           &                                                                         &                        &                                                                                  & Daytime & Nighttime & Sunset & Cloudy & Foggy & Rainy        \\ 
\hline
MVSEC \cite{zhu2018multivehicle}             & 2018                  & R                                                                         & \ding{52}                                                                       & \ding{52}                     & \ding{56}                                                                                & \ding{52}      & \ding{52}        & \ding{56}      & \ding{52}    & \ding{56}   & \ding{56}          \\
DSEC \cite{gehrig2021dsec}                     & 2020                 & R                                                                         & \ding{52}                                                                      & \ding{52}                     & \ding{56}                                                                              & \ding{52}      & \ding{52}        & \ding{52}       & \ding{52}    & \ding{56}   & \ding{56}          \\
Ekubric \cite{wan2023rpeflow}            & 2023                  & S                                                                         & \ding{52}                                                                      & \ding{52}                     & \ding{56}                                                                              & \ding{52}      & \ding{56}       & \ding{56}      & \ding{52}    & \ding{56}   & \ding{56}          \\
KITTI-Event \cite{zhou2024bring}       & 2023                  & R+S                                                                       & \ding{52}                                                                      & \ding{52}                     & \ding{56}                                                                              & \ding{52}      & \ding{56}       & \ding{56}      & \ding{52}    & \ding{56}   & \ding{56}          \\
FlyingThings-Event \cite{wan2023rpeflow}         & 2023                  & S                                                                       & \ding{52}                                                                      & \ding{52}                     & \ding{56}                                                                              & \ding{52}      & \ding{56}       & \ding{56}      & \ding{52}    & \ding{56}   & \ding{56}          \\
BlinkVision \cite{li2024blinkvision}          & 2024                  & S                                                                         & \ding{52}                                                                      & \ding{52}                     & \ding{56}                                                                              & \ding{52}      & \ding{56}       & \ding{56}      & \ding{52}    & \ding{56}   & \ding{56}          \\
\rowcolor{mygray}
CarlaEvent3d              & 2024                  & S                                                                         & \ding{52}                                                                      & \ding{52}                     & \ding{52}                                                                               & \ding{52}      & \ding{52}        & \ding{52}       & \ding{52}    & \ding{52}    & \ding{52}           \\
\hline\hline
\end{tabular}

% \vspace{-4mm}
\caption{Related motion estimation dataset based on event camera.}
\label{table:dataset}
\end{table*}

% \begin{table}[h] 
% \small
% % \vspace{-3mm}
% \begin{center} 
%     \renewcommand{\arraystretch}{1.}
%     \renewcommand{\tabcolsep}{6pt}
%     % \centering
%     % \small
%     \begin{tabular}{c|cc|c}
%     \Xhline{1.\arrayrulewidth}
%     \hline 
%      \multirow{2}{*}{Method}& \multicolumn{2}{c|}{UCF-101} & FaceForensics\\ 
%      \cline{2-4}
%      & FVD$\downarrow$  & IS$\uparrow$  & FVD$\downarrow$ \\ 
%     \Xhline{1.\arrayrulewidth}
%     \hline  
%     CV-VAE & 587.9 & 84.7 & 328.2 \\  
%     OS-VAE & 674.1 & 85.2 & 316.3 \\
%     OD-VAE &  565.2 & 82.5 & 285.5 \\
%     \rowcolor{mygray}
%     IV-VAE & \textbf{557.5} & \textbf{85.7} & \textbf{259.8}  \\
%     \hline
%     \Xhline{1.\arrayrulewidth}
%     \end{tabular}
    
%     \end{center}
%     \vspace{-4mm} 
%     \caption{\textbf{Video generation results.}}
%     \label{table:gen}
%     \vspace{-2mm} 
% \end{table}

\section{Data Collection Process}
We employ the Carla simulator \cite{dosovitskiy2017carla} to generate the 3D motion dataset \textbf{CarlaEvent3d} in a driving environment. Carla provides realistic simulations of various weather conditions, as well as optical flow on the camera plane and relative depth labels during the driving process. For event generation, we first leverage the UE4 engine in Carla to produce high-frame-rate video and then simulate events using the DVS Voltemter algorithm \cite{lin2022dvs} integrated into the Carla simulation workflow. The detailed data generation process is illustrated in \cref{fig:simulation}. Ultimately, we obtained 75 sequences across diverse environments—including rain, fog, and night scenes—resulting in a total of 22,125 event-image-3D motion labels tuples.

\subsection{Forward Optical Flow Generation}
Generating optical flow via the Carla simulator presents two major challenges: first, the precision of the optical flow is limited to 10 digits in each direction; second, the simulator produces backward optical flow rather than the forward optical flow that is commonly used. This precision limitation arises because Carla employs the Emissive Color property of UE4 materials to output optical flow, yielding up to three channels of \texttt{float16} color with each channel providing up to 10 bits of valid information. To enhance accuracy, we utilize two independent materials to encode the horizontal and vertical components of the optical flow. Each material continues to use Emissive Color, but the first ten bits and the last ten bits of the optical flow are encoded into separate channels, thereby improving the final output quality. For forward optical flow generation, we adopt the efficient Inverse Optical Flow algorithm \cite{sanchez2013efficient} and combine it with the depth map to transform the backward optical flow into forward optical flow. Moreover, since Carla provides instance segmentation results, we further refined the algorithm so that the inverse computation of the matching pixel position accounts for both depth approximation and semantic label consistency, ultimately yielding more accurate forward optical flow estimates.

\subsection{Motion in Depth Generation}
The depth motion label is computed by warping the depth value from the target moment to the initial moment using the forward optical flow:
\begin{align}
    \mathcal{M} = \frac{Z_1(x+v(x))}{Z_0(x)},
\end{align}
where \(Z_1\) is the target moment depth map, \(Z_0\) is the initial moment depth map, and \(v\) represents the forward optical flow. To address instability and uncertainty in depth labeling at object boundaries, we mask out estimation results near these boundaries.

After obtaining the forward optical flow and motion in depth, we combine the initial depth value with the camera's internal parameters to derive the scene flow.

\section{Comprehensive Experimental Results}

In this section, we present additional experimental results for 3D motion estimation.

\begin{figure}
    \centering
    \begin{subfigure}[b]{\linewidth}
         \centering
         \includegraphics[width=1.\linewidth]{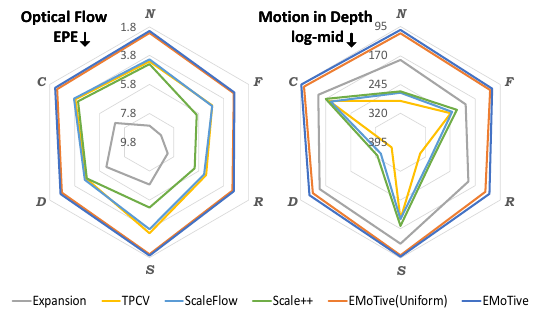}
         \caption{Evaluation of Motion estimation performance in different environments. \textbf{\textit{N}}:Nighttime, \textbf{\textit{D}}:Daytime, \textbf{\textit{S}}:Sunset, \textbf{\textit{C}}:Cloudy, \textbf{\textit{F}}:Foggy, \textbf{\textit{R}}:Rainy.}
         \label{fig:scenes}
     \end{subfigure}
     \begin{subfigure}[b]{.75\linewidth}
         \centering
         \includegraphics[width=1.\linewidth]{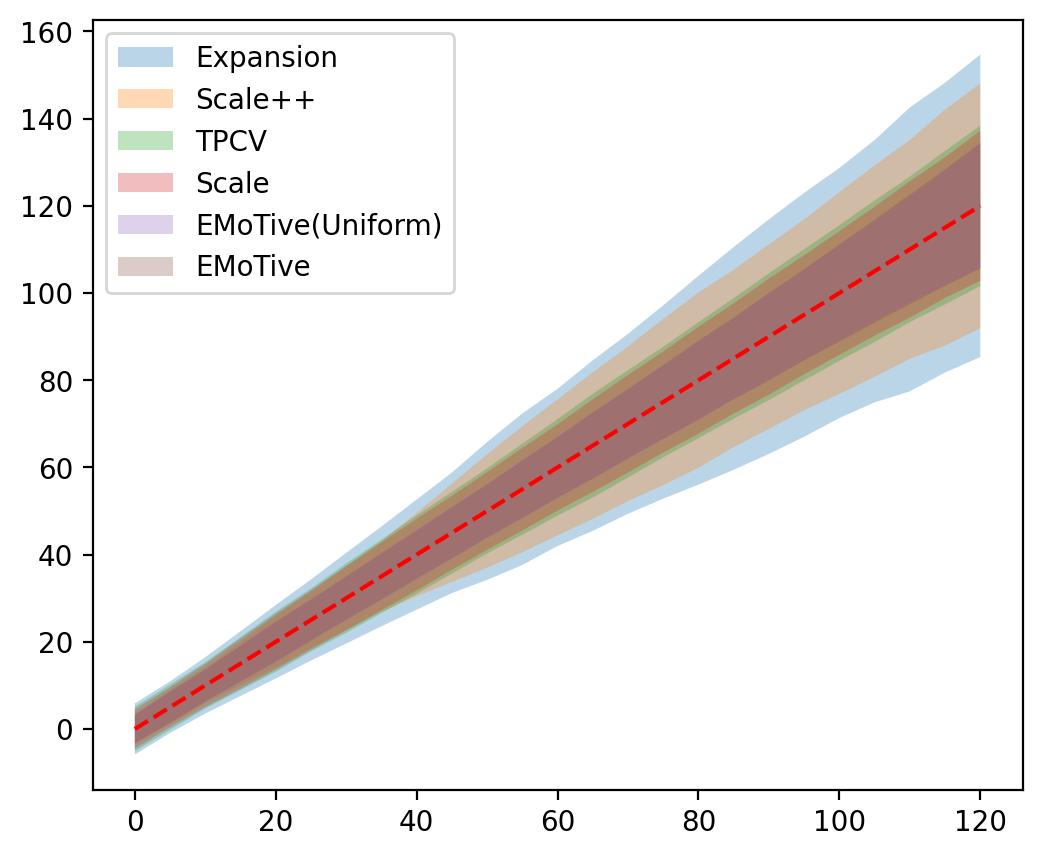}
         \caption{The plot for motion amplitude estimation capability}
         \label{fig:speeds}
     \end{subfigure}
    \caption{Verification of Motion estimation ability}
\end{figure}

\subsection{Performance in Different Scenes}
We evaluated our method across various scenes in CarlaEvent3D, including rain, fog, night, cloudy daytime, sunny, and dusk, as shown in \cref{fig:scenes}. Even under challenging conditions—such as low-light noise, insufficient contrast, and rainwater interference—EMoTive achieved excellent motion estimation, demonstrating robust algorithmic stability. Notably, in foggy conditions, methods like Expansion experienced a significant decline in optical flow estimation (a decrease of 2.48 px in accuracy), whereas EMoTive exhibited only a 1.01 px change, thereby maintaining a higher degree of accuracy.

\subsection{Performance of Motion Amplitude Estimation}
To assess the capability of motion amplitude estimation, we evaluated the methods over a range of amplitudes from 0 to 120 px/100ms. Accuracy was quantified by the standard deviation within different motion intervals; a smaller standard deviation indicates a closer approximation to the true motion and, hence, better estimation accuracy. The validation results are presented in \cref{fig:speeds}. The findings reveal that as the motion amplitude increases, the performance of all methods deteriorates, underscoring the significant challenge posed by fast motion. However, the proposed EMoTive method maintains lower standard deviations at high speeds and exhibits a relatively consistent error range across different motion intervals. This outcome indicates that the event-guided non-uniform trajectory design provides robust tracking performance across various motion intensities, thereby enabling stable motion estimation.

\section{Result Visualization on CarlaEvent3D}
\begin{figure*}[tp]
     \centering
     \includegraphics[width=0.85\linewidth]{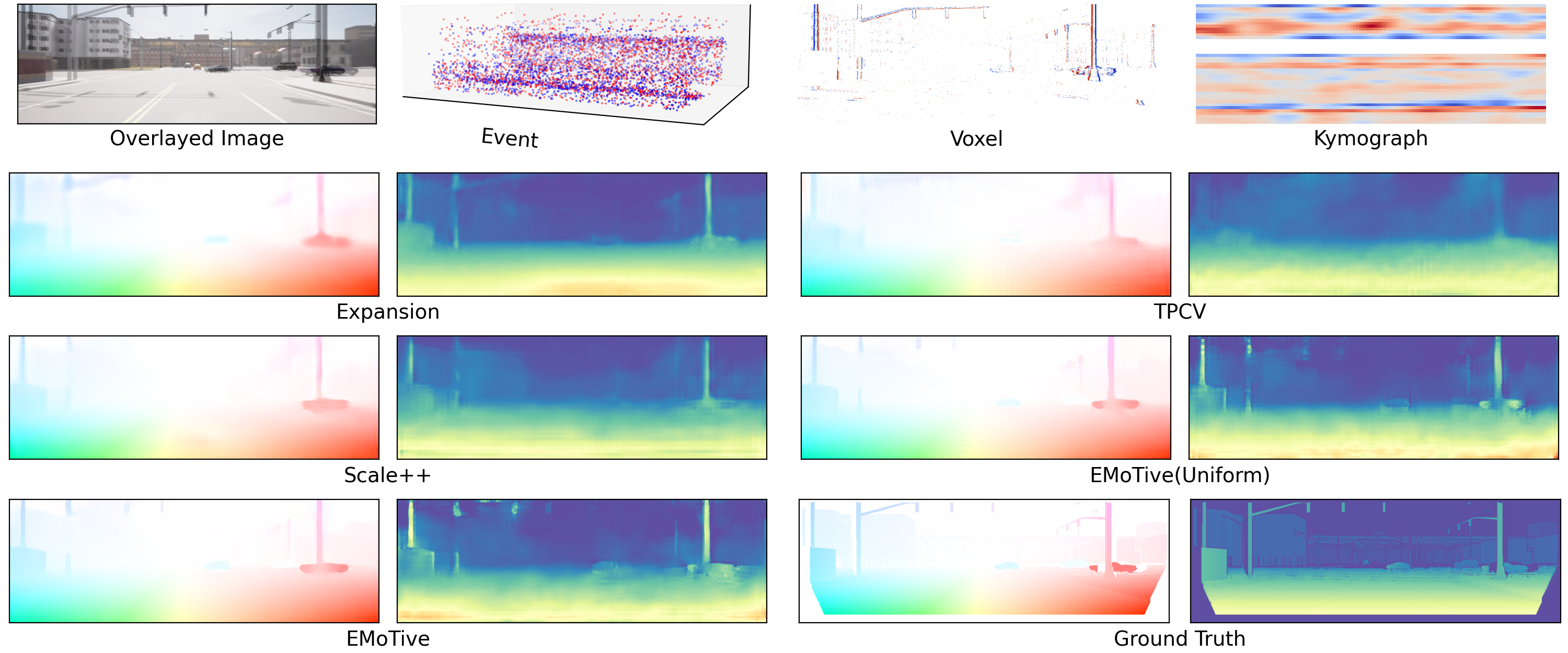}
     \caption{Qualitative comparison on the CarlaEvent3D dataset (daytime). The left is optical flow and the right is motion in depth estimation.}
     \label{fig:daytime}
\end{figure*}

\begin{figure*}[tp]
     \centering
     \includegraphics[width=0.85\linewidth]{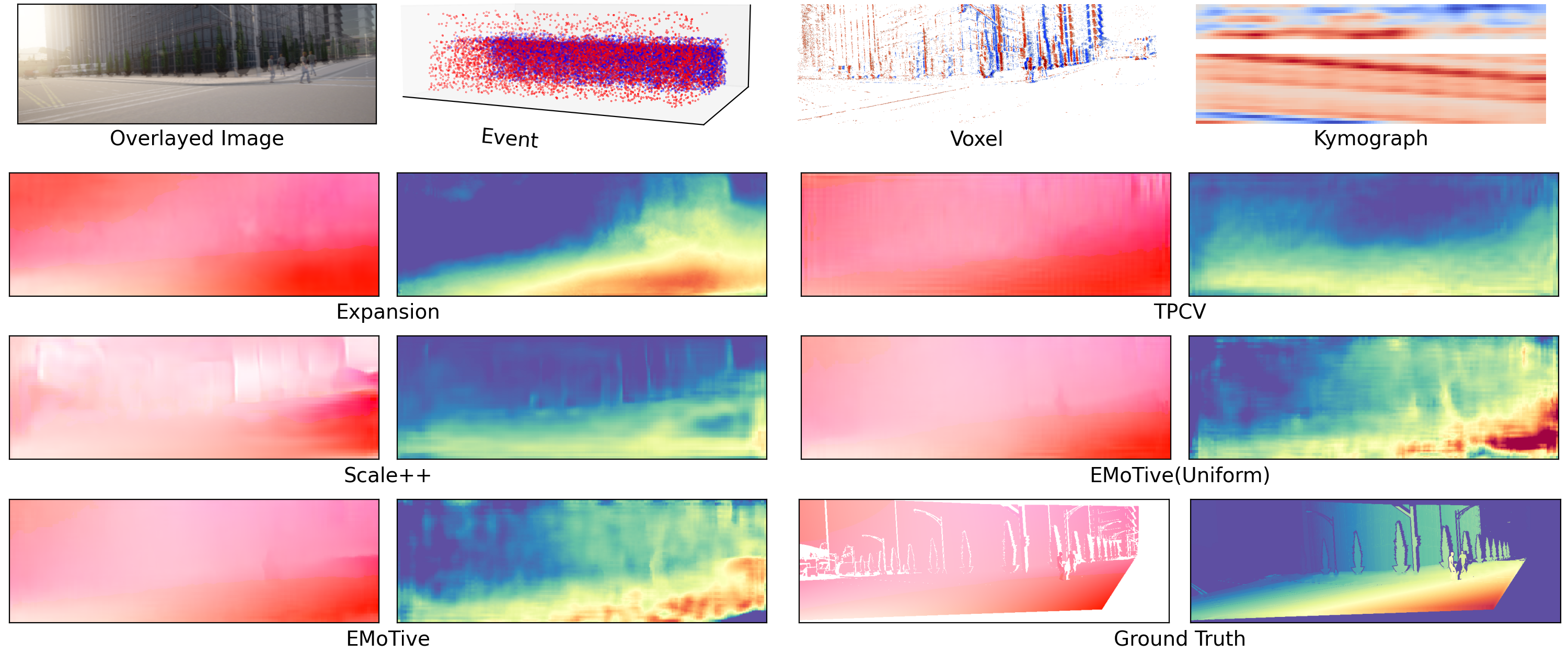}
     \caption{Qualitative comparison on the CarlaEvent3D dataset (sunset).}
     \label{fig:sunset}
\end{figure*}

\begin{figure*}[tp]
     \centering
     \includegraphics[width=0.85\linewidth]{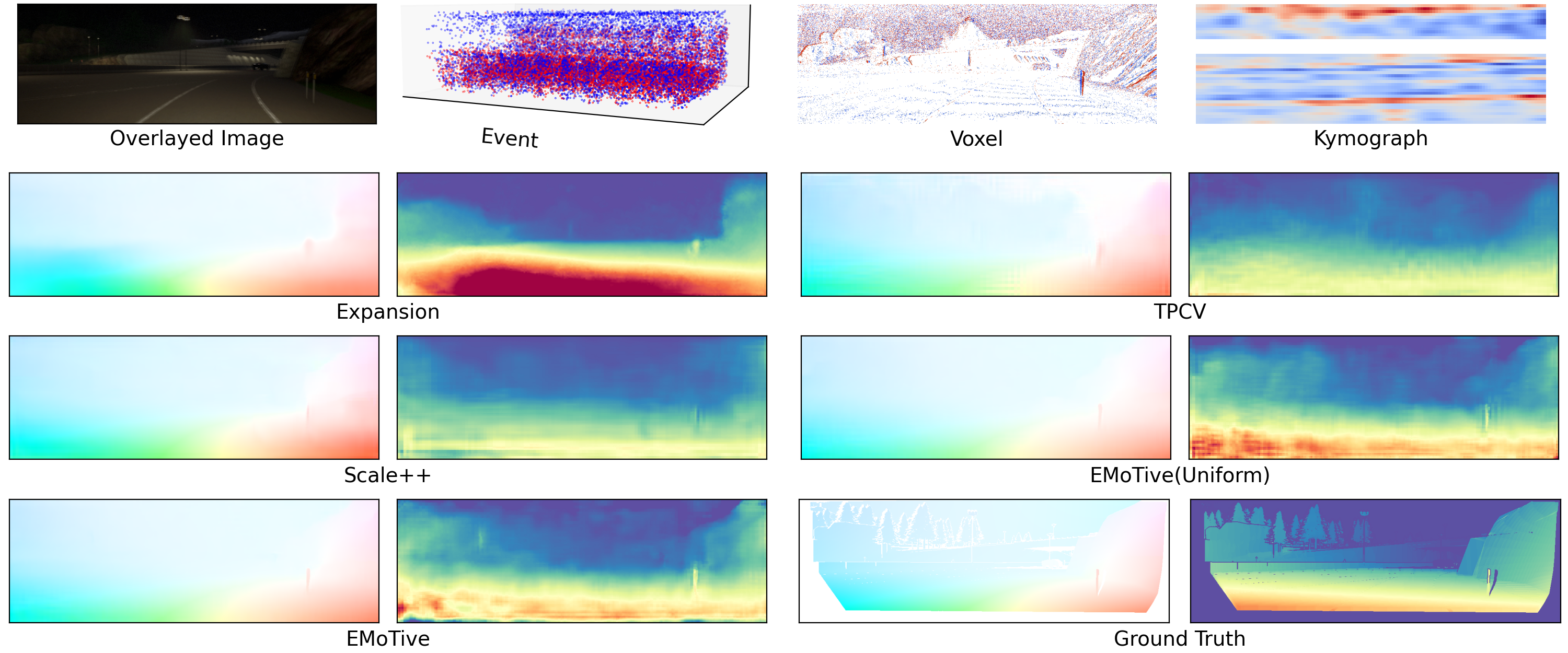}
     \caption{Qualitative comparison on the CarlaEvent3D dataset (nighttime).}
     \label{fig:nighttime}
\end{figure*}

\begin{figure*}[tp]
     \centering
     \includegraphics[width=0.85\linewidth]{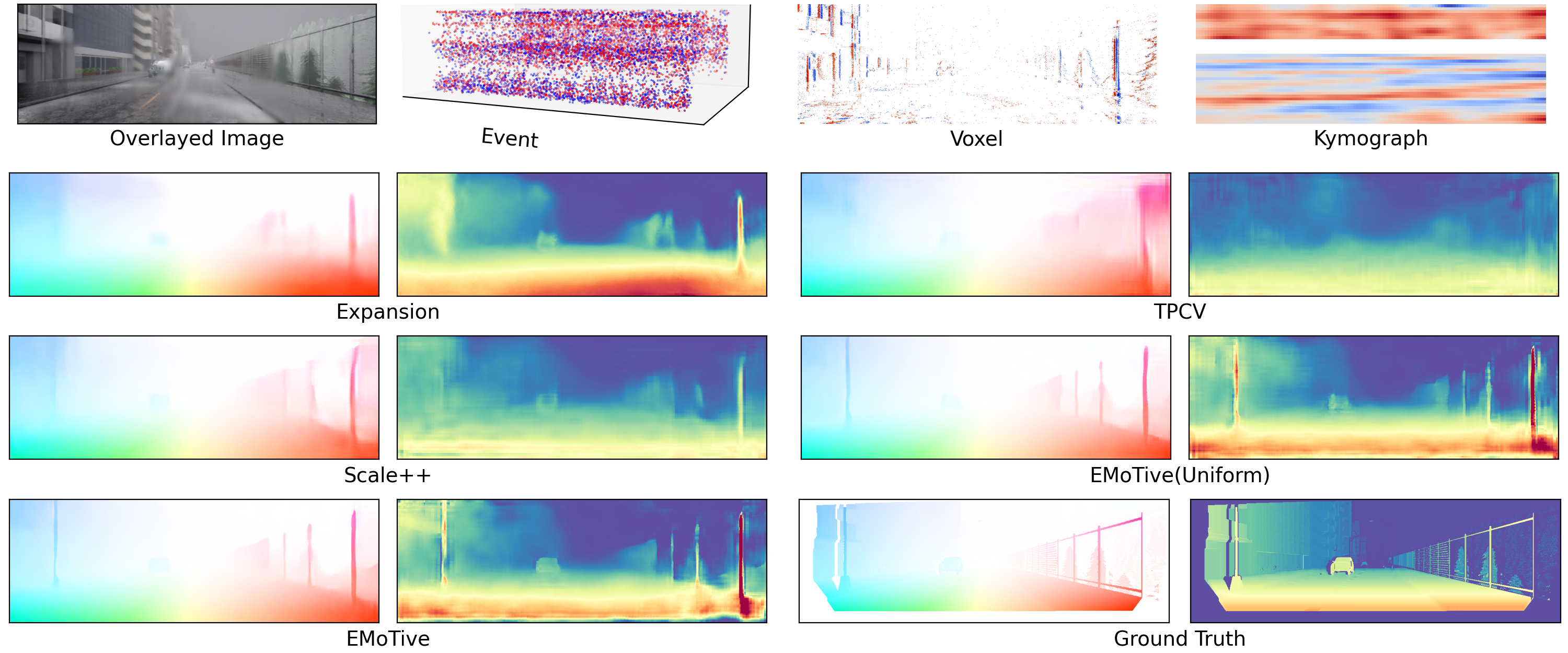}
     \caption{Qualitative comparison on the CarlaEvent3D dataset (rainy).}
     \label{fig:rain}
\end{figure*}

\begin{figure*}[tp]
     \centering
     \includegraphics[width=0.85\linewidth]{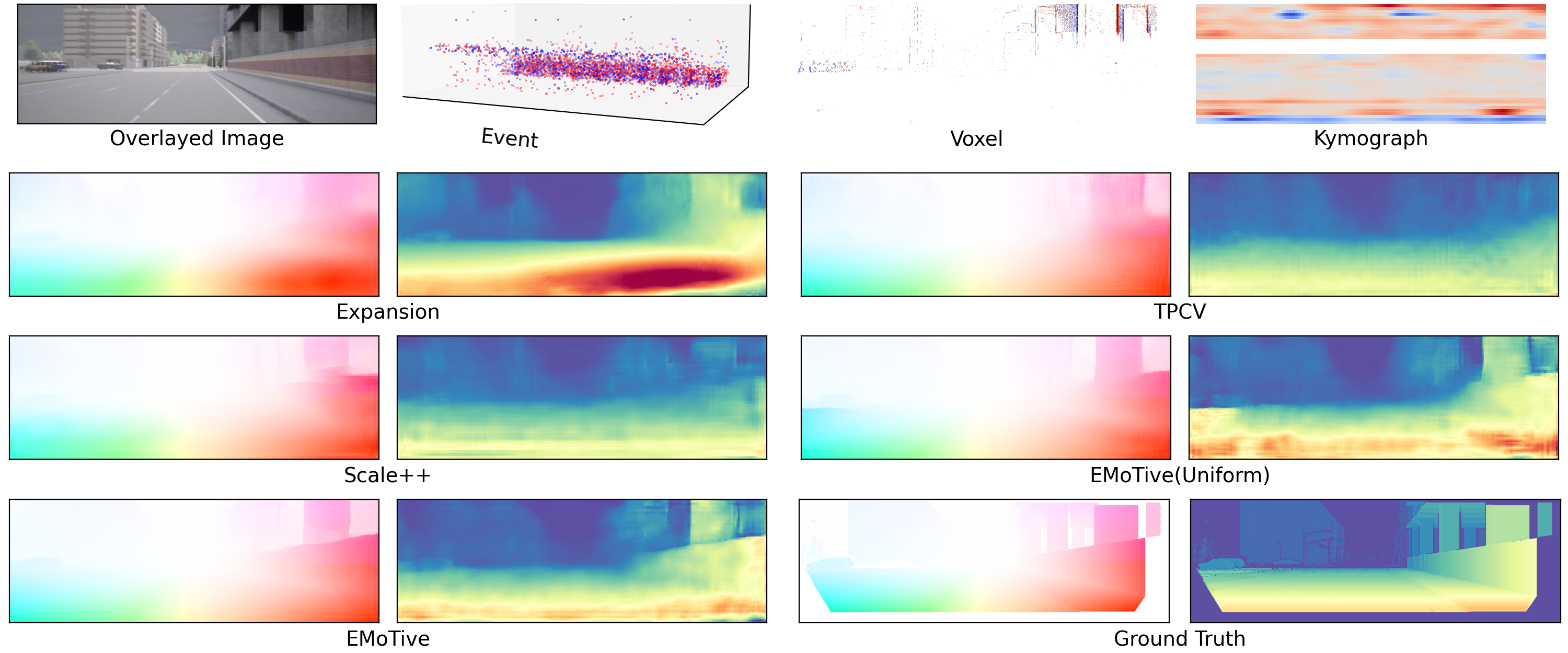}
     \caption{Qualitative comparison on the CarlaEvent3D dataset (cloudy).}
     \label{fig:cloudy}
\end{figure*}

\begin{figure*}[tp]
     \centering
     \includegraphics[width=0.85\linewidth]{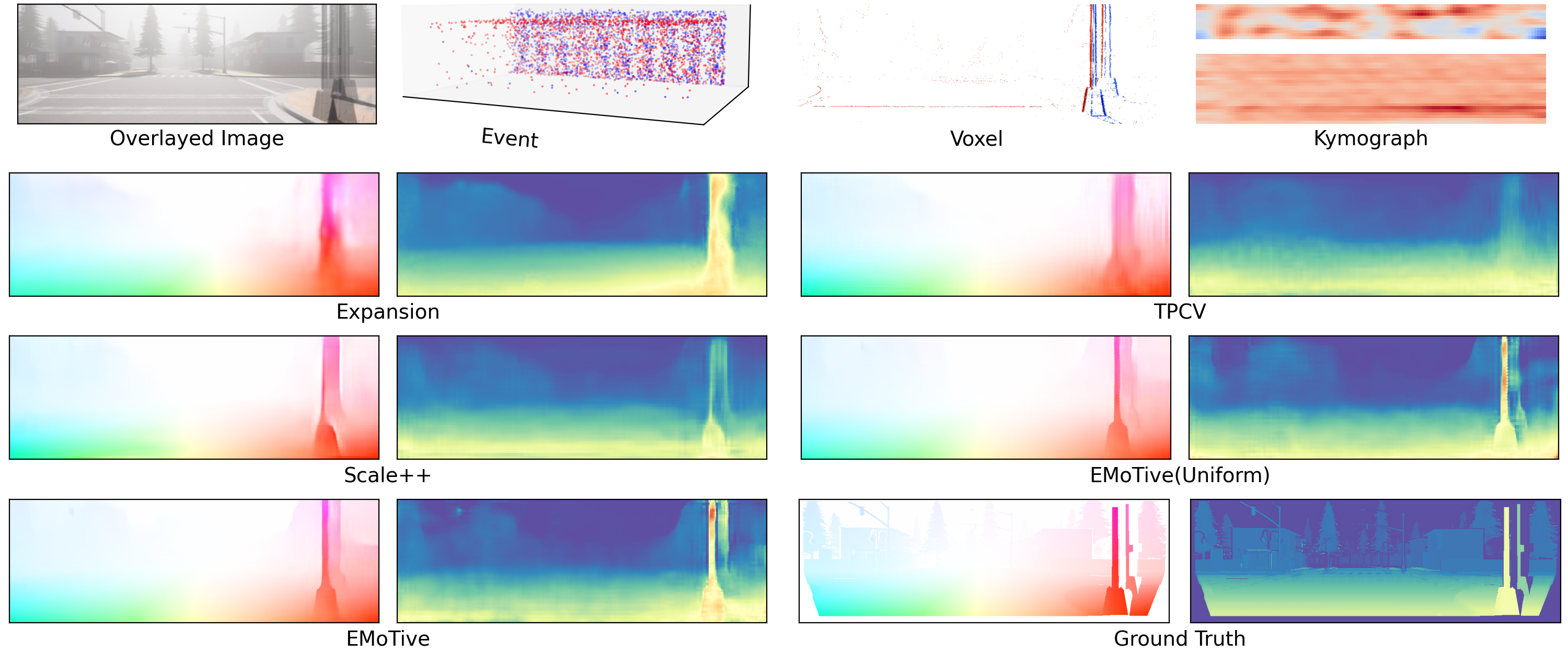}
     \caption{Qualitative comparison on the CarlaEvent3D dataset (foggy).}
     \label{fig:foggy}
\end{figure*}

To illustrate the motion estimation performance, we present additional visualizations on the CarlaEvent3D dataset across various scenes. These visualizations include images at the start and end moments, the event sequence input, event voxel projection, event kymograph projection, and the motion estimation outputs from several methods, including our proposed EMoTive model (see \cref{fig:daytime,fig:sunset,fig:nighttime,fig:rain,fig:cloudy,fig:foggy}). The results demonstrate that EMoTive achieves clearer spatial boundaries in both optical flow and depth motion estimation. Moreover, for fast-moving objects, its 3D motion estimation is more comprehensive and accurate. In particular, the non-uniform curve representation in EMoTive yields a motion estimation distribution that more closely aligns with the ground truth, resulting in a smoother depiction of motion.

\end{document}